\documentclass[fleqn,10pt]{wlpeerj}

\usepackage{algorithm}
\usepackage{algpseudocode}
\usepackage{algcompatible}
\usepackage{float}
\usepackage{microtype}
\usepackage{graphicx}
\usepackage{subfigure}
\usepackage{booktabs} 
\usepackage{url}
\usepackage[hidelinks]{hyperref}
\hypersetup{
  colorlinks   = true, 
  urlcolor     = blue, 
  linkcolor    = black, 
  citecolor    = blue   
}

\title{Positive Unlabeled Learning Selected Not At Random \mbox{(PULSNAR):} class proportion estimation when the SCAR assumption does not hold}

\author[1]{Praveen Kumar}
\author[1]{Christophe G. Lambert}
\affil[1]{University of New Mexico, Department of Internal Medicine, Division of Translational Informatics, Albuquerque, NM, USA}
\corrauthor[1]{Christophe G. Lambert}{cglambert@salud.unm.edu}

\begin{abstract}
Positive and Unlabeled (PU) learning is a type of semi-supervised binary classification where the machine learning algorithm differentiates between a set of positive instances (labeled) and a set of both positive and negative instances (unlabeled). PU learning has broad applications in settings where confirmed negatives are unavailable or difficult to obtain, and there is value in discovering positives among the unlabeled (e.g., viable drugs among untested compounds). Most PU learning algorithms make the \emph{selected completely at random} (SCAR) assumption, namely that positives are selected independently of their features. However, in many real-world applications, such as healthcare, positives are not SCAR (e.g., severe cases are more likely to be diagnosed), leading to a poor estimate of the proportion, $\alpha$, of positives among unlabeled examples and poor model calibration, resulting in an uncertain decision threshold for selecting positives. PU learning algorithms vary; some estimate only the proportion, $\alpha$, of positives in the unlabeled set, while others calculate the probability that each specific unlabeled instance is positive, and some can do both. We propose two PU learning algorithms to estimate $\alpha$, calculate calibrated probabilities for PU instances, and improve classification metrics: i) PULSCAR (positive unlabeled learning selected completely at random), and ii) PULSNAR (positive unlabeled learning selected not at random). PULSNAR employs a divide-and-conquer approach to cluster SNAR positives into subtypes and estimates $\alpha$ for each subtype by applying PULSCAR to positives from each cluster and all unlabeled. In our experiments, PULSNAR outperformed state-of-the-art approaches on both synthetic and real-world benchmark datasets.
\end{abstract}

\begin{document}

\flushbottom
\maketitle
\thispagestyle{empty}

\section*{Introduction}

In a standard binary supervised classification problem, the classifier (e.g., decision trees, support vector machines, etc.) is given training instances $\mathcal{X}$ with features $x$ and their labels $y=0$ (negative) or $y=1$ (positive). The classifier learns a model $f:  \mathcal{X} \rightarrow 0, 1$, which classifies an unlabeled instance as positive or negative based on $x$. It is often challenging, expensive, and even impossible to annotate large datasets in real-world applications \citep{pulsnar_1}, and frequently only positive instances are labeled. Unlabeled instances with their features can be classified via positive and unlabeled (PU) learning \citep{pulsnar_1, pulsnar_2}. Some of the PU learning literature focuses on improving classification metrics, and others focus on the problem of estimating the fraction, $\alpha$, of positives among the unlabeled instances. Although this work focuses on the latter, calibration and enhancing classification performance are also addressed.

PU learning problems abound in many domains \citep{pulsnar_50}. For instance, in electronic healthcare records, the lack of a diagnosis code does not confirm a patient's negative disease status, as negatives are not routinely recorded nor have billing codes, making traditional supervised learning impractical. Much medical literature is dedicated to estimating disease incidence and prevalence but contends with incomplete medical assessment and recording. The potential to assess disease incidence without costly in-person assessment or chart reviews could have substantial public health benefits. In market research, one typically has a modest set of positives, say of customers or buyers of a product, has a set of attributes over both the positives and a large population of unlabeled people of size $N$, and wishes to establish the size of the addressable market, $\alpha N$.

The majority of PU learning algorithms use the \emph{selected completely at random} (SCAR) assumption, which states that the labeled positive examples are randomly selected from the universe of positives. That is, the labeling probability of any positive instance is constant \citep{pulsnar_2}. This assumption may fail in real-world applications. For example, in email spam detection, positive instances labeled from an earlier time period could differ from later spam due to adaptive adversaries. 

Although some PU learning algorithms have shown promising performance on different machine learning (ML) benchmark SCAR datasets, the development of PU learning algorithms to estimate the extent of undercoding in large and highly imbalanced \emph{selected not at random} (SNAR) real-world data remains an active research area. Class imbalance in a PU setting generally means the number of unlabeled instances is large compared to the labeled positive examples. Also, current PU learning approaches have rarely explored how to calculate well-calibrated probabilities for PU examples in SCAR and SNAR settings. In addition, few PU algorithms have been assessed when $\alpha$ is small ($\leq 5\%$), where performance is expected to suffer.

In this paper, we propose a PU learning approach to estimate $\alpha$ when positives are SCAR or SNAR, and evaluating its performance in simulated and real data. We assess the performance with class imbalance in both modest and large datasets and over a rigorous $\alpha$ range. Our contributions are summarized as follows:
\begin{enumerate}
    \itemsep0em
    \item We propose PULSCAR, a PU learning algorithm for estimating $\alpha$ when the SCAR assumption holds. It uses kernel density estimates of the positive and unlabeled distributions of ML probabilities to estimate $\alpha$. The algorithm employs the beta distribution to estimate density and introduces an objective function whose derivative maximum provides a rapid, robust estimate of $\alpha$.
    \item We propose PULSNAR, a PU learning algorithm for estimating $\alpha$ when the positives are SNAR. It employs a clustering approach to group SNAR positives into subtypes and estimates separate $\alpha$ for each subtype using PULSCAR on positives from each cluster and all unlabeled instances. The final $\alpha$ is calculated by aggregating the $\alpha$ estimated for each subtype.
    \item We propose methods to calibrate the probabilities of PU examples to their true (unknown) labels and improve the classification performance in SCAR and SNAR settings.
\end{enumerate}

\section*{Related work}
\label{sectionRelatedWork}
Early PU learning methods \citep{pulsnar_12, pulsnar_13, pulsnar_14} generally followed a two-step heuristic: i) identify strong negative examples from the unlabeled set, and then ii) apply an ML algorithm to given positive and identified negative examples. In contrast, \citep{pulsnar_15} extracted high-quality positive and negative examples from the unlabeled set and then applied classifiers to those data. Some recent work iteratively identifies better negatives \citep{pulsnar_47}, or combines negative-unlabeled learning with unlabeled-unlabeled learning \citep{pulsnar_44}. 

Studies predominantly centered around the SCAR assumption focused on estimating the proportion of positives among the unlabeled examples with or without the PU classifier. \citep{pulsnar_2} introduced the SCAR assumption and proposed a PU method to estimate the mixture proportion under the SCAR assumption. By partially matching the class-conditional density of the positive class to the input density under Pearson divergence minimization, \citep{pulsnar_20} estimated the mixture coefficient. \citep{pulsnar_21} proposed a nonparametric class prior estimation technique, AlphaMax, using two-component mixture models. The kernel embedding approaches KM1 and KM2 \citep{pulsnar_22} showed that the algorithm for mixture proportion estimation converges to the true prior under certain assumptions. Estimating the class prior through decision tree induction (TICE) \citep{pulsnar_23} provides a lower bound for label frequency under the SCAR assumption. Using the SCAR assumption, DEDPUL \citep{pulsnar_24} estimates $\alpha$ by applying a compute-intensive EM-algorithm to probability densities; the method also returns uncalibrated probabilities.

Other studies focused on learning a classifier from PU data. \citep{pulsnar_16} converts PU data learning into a noisy learning problem by designating all unlabeled instances as negatives. They employ a linear function to learn from these noisy examples using weighted logistic regression. Confident learning (CL) \citep{pulsnar_25} combines the principle of pruning noisy data, probabilistic thresholds to estimate noise, and sample ranking. Multi-Positive and Unlabeled Learning \citep{pulsnar_26} extends PU learning to multi-class labels. Oversampling the minority class \citep{pulsnar_27,pulsnar_28} or undersampling the majority class are not well-suited approaches for PU data due to contamination in the unlabeled set; \citep{pulsnar_31} uses a re-weighting strategy for imbalanced PU learning. 

Recent studies have focused on labeling/selection bias to address the SCAR assumption not holding. \citep{pulsnar_42, pulsnar_43} used propensity scores to address labeling bias and improve classification. Using the propensity score, based on a subset of features, as the labeling probability for positive examples, \citep{pulsnar_42} reduced the Selected At Random (SAR) problem into the SCAR problem to learn a classification model in the PU setting. The ``Labeling Bias Estimation'' approach was proposed by \citep{pulsnar_48} to label the data by establishing the relationship among the feature variables, ground-truth labels, and labeling conditions.

\section*{Problem Formulation and Algorithms}
\label{sectionProblemFormulation}
In this section, we explain: i) the SCAR and SNAR assumptions, ii) our PULSCAR algorithm for SCAR data and PULSNAR algorithm for SNAR data, iii) kernel density and bandwidth estimation techniques, and iv) method to find the number of clusters in the labeled positive set. Our method to calibrate probabilities and enhance classification performance using PULSCAR/PULSNAR is in Appendix \ref{appendix1} and \ref{appendix2}, respectively.  

\subsection*{SCAR assumption and SNAR assumption}
In PU learning settings, a positive or unlabeled example can be represented as a triplet ($x, y, s$) where ``$x$'' is a vector of the attributes, ``$y$'' the actual class, and ``$s$'' a binary variable representing whether or not the example is labeled. If an example is labeled ($s=1$), it belongs to the positive class ($y=1$) i.e., $p(y=1|s=1)=1$. If an example is not labeled ($s=0$), it can belong to either class. Since only positive examples are labeled, $p(s=1|x, y=0)=0$ \citep{pulsnar_2}. Under the SCAR assumption, a labeled positive is an independent and identically distributed (i.i.d) example from the positive distribution, i.e., positives are selected independently of their attributes. Therefore, $p(s=1|x, y=1) = p(s=1|y=1)$ \citep{pulsnar_2}.

For a given dataset, $p(s=1|y=1)$ is a constant and is the fraction of labeled positives. If $|P|$ is the number of labeled positives, $|U|$ is the number of unlabeled examples, and $\alpha$ is the unknown fraction of positives in the unlabeled set, then 
\begin{gather}
p(s=1) = \frac{|P|}{|P|+|U|} \quad \text{and} \quad p(y=1) = \frac{|P|+\alpha |U|}{|P|+|U|} \quad \text{(class prior)} \nonumber \\
p(s=1|y=1) = \frac{p(y=1|s=1)p(s=1)}{p(y=1)} = \frac{p(s=1)}{p(y=1)} \quad \text{, since $p(y=1|s=1) = 1$} \nonumber\\
= \frac{|P|}{|P|+\alpha |U|}  \text{, which is a constant.}
\end{gather}

On the contrary, under the SNAR assumption, the probability that a positive example is labeled is not independent of its attributes. Stated formally, the assumption is that
$p(s=1|x, y=1) \neq p(s=1|y=1)$ i.e. $p(s=1|x, y=1)$ is not a constant, which can be proved by Bayes' rule (Appendix \ref{appendix0}). 

The SCAR assumption can hold when both labeled and unlabeled positives: a) are not subclass mixtures, sharing similar attributes; b) belong to \emph{k} subclasses ($1 \dots k$), with equal subclass proportions in both positive and unlabeled sets. Intra-subclass examples will have similar attributes, whereas the inter-subclass examples may not have similar attributes. E.g., in patients positive for diabetes, type 1 patients will be in one subclass, and type 2 patients will be in another. The SCAR assumption can fail when labeled and unlabeled positives are from \emph{k} subclasses, and the proportion of those subclasses is different in positive and unlabeled sets. Suppose both positive and unlabeled sets have subclass 1 and subclass 2 positives, and in the positive set, their ratio is 30:70. If the ratio is also 30:70 in the unlabeled set, the SCAR assumption will hold. If it was different, say 80:20, the SCAR assumption would not hold.

\subsubsection*{PU data assumptions}
Our proposed PU algorithms rely on the following assumptions about the PU data: $1)$ Positive examples have correct labels, meaning no negative example is marked as positive. Only the unlabeled set contains a mixture of positive and negative instances. $2)$ Unlabeled positive instances have analogous counterparts (examples with similar features) in the labeled positive set. If the second assumption does not hold, our approaches may underestimate $\alpha$. 

Our PU algorithms involve running a machine learning model on a combined set of positive and unlabeled instances. The goal is to obtain machine learning-predicted probabilities for all instances to determine $\alpha$, the fraction of positives among the unlabeled examples. It is worth noting that in our PU algorithms, $\alpha$ is not the same as the class prior. The class prior represents the proportion of positives, labeled and unlabeled combined, in the dataset and can be estimated if $\alpha$ is known.

\subsection*{Positive and Unlabeled Learning Selected Completely At Random (PULSCAR) Algorithm}

Given any ML algorithm, $\mathcal{A}(x)$, that generates [0\dots1] probabilities for the data based on covariates $x$, let $f_p(x)$, $f_n(x)$, and $f_u(x)$ be probability density functions (PDFs) corresponding to the probability distribution of positives, negatives, and unlabeled respectively. Let $\alpha$ be the unknown proportion of positives in the unlabeled, then 

\begin{align}\label{obj_func_prop}
 &f_u(x) = \alpha f_p(x) + (1-\alpha) f_n(x) \text{,}\quad \text{ using the law of total probability} && \nonumber \\ 
 &\Rightarrow f_u(x) \geq \alpha f_p(x) \text{,}\quad \text{ since } 0  \leq \alpha \leq 1 \text{ and }  f_n(x) \text{ is non-negative} && \nonumber \\ 
 &\Rightarrow 0 \leq \alpha f_p(x) \leq f_u(x) \text{,}\quad \text{ since } f_p(x) \text{ is non-negative}
\end{align} 

From property \ref{obj_func_prop}, a key observation is that $\alpha f_p(x)$ should not exceed $f_u(x)$ anywhere, ensuring that the $\alpha$ is bounded ($0 \leq \alpha \leq \frac{f_u(x)}{f_p(x)}$). 

PULSCAR estimates $\alpha$ by finding the value $\alpha$ where the following objective function maximally changes:

\begin{equation}\label{eq:obj_func}
    f(\alpha) = log(|\min(f_u(x) - \alpha f_p(x))|+ \epsilon) \text{, where } 
 \epsilon = |\min(f_p(x))| \text{ if } \min(f_p(x)) \neq 0 \text{, else } \epsilon = 10^{-10}
\end{equation}

\begin{figure}[H]
\centering
\includegraphics[width=0.95\columnwidth]{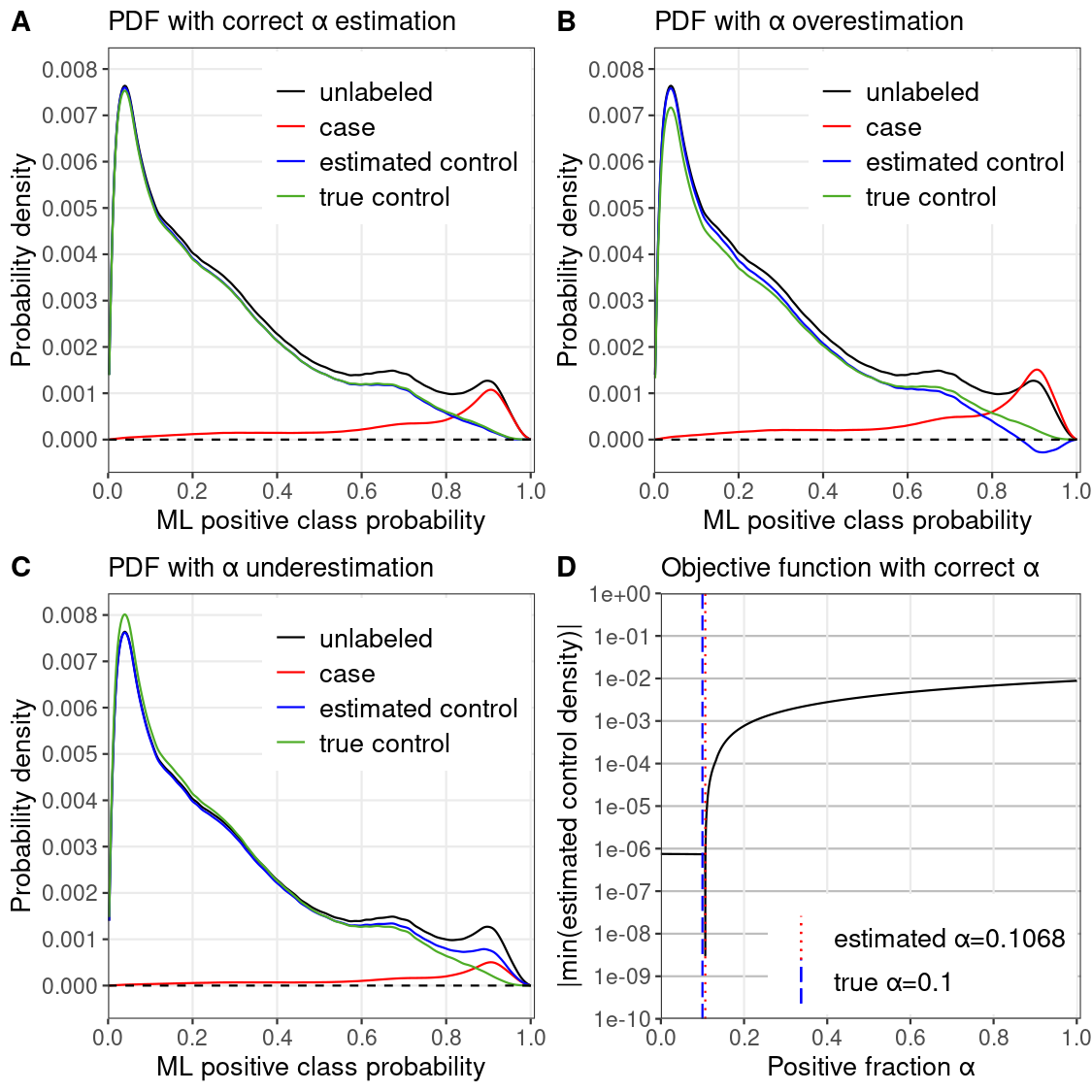} 
\caption{\textbf{PULSCAR algorithm visual intuition}. PULSCAR finds the smallest $\alpha$ such that $f_u(x) - \alpha  f_p(x)$ is everywhere positive in [0 \dots 1]. A) Kernel density estimates for simulated data with $\alpha=10\%$ positives in the unlabeled set -- estimated negative density (blue) nearly equals the ground truth (green). B) Overweighting the positive density by $\alpha=15\%$ results in the estimated negative density (blue), $f_u(x) - \alpha f_p(x)$ dropping below zero. C) Underweighting the positive density by $\alpha=5\%$ results in the estimated negative density (blue) being higher than the ground truth (green). D) Objective function with estimated $\alpha=10.68\%$ selected where the finite-differences estimate of the slope is largest -- very close to ground truth $\alpha=10\%$.}
\label{fig:densityerrorplot}
\end{figure}

Property \ref{obj_func_prop}, $\alpha f_p(x) \leq f_u(x)$, guided the reasoning behind the design choice of the objective function. The intuition behind the objective function is that $|\min(f_u(x) - \alpha f_p(x))|$ approaches zero at the point where $\alpha f_p(x)$ equals $f_u(x)$, see Figure \ref{fig:densityerrorplot}D. When we take the logarithm of $|\min(f_u(x) - \alpha f_p(x))|$, the resulting value tends toward $-\infty$ as $|\min(f_u(x) - \alpha f_p(x))|$ approaches zero. Consequently, when $|\min(f_u(x) - \alpha f_p(x))|$ is not zero, there is a maximum change in the value of $\log(|\min(f_u(x) - \alpha f_p(x))|)$ (from $-\infty$ to some value). That is why we locate the point where $\alpha f_p(x)$ equals $f_u(x)$ to place an upper bound on $\alpha$. The reason we use the logarithm is it steeply approaches $-\infty$ as $|\min(f_u(x) - \alpha f_p(x))|$ approaches zero. Adding $\epsilon$ to prevent $log(0)$ and using finite differences to find the max change in slope gives a robust estimator that is resilient to noise. The objective function may not be convex; if multiple points with the same maximal change occur, we take the one closest to zero as the $\alpha$ estimate. This approach eliminates the need for implementing an iterative solver technique, accounting in part for the speed of our algorithm. 

We use beta kernel density estimates on ML-predicted class 1 probabilities of positives and unlabeled to estimate $f_p(x)$ and $f_u(x)$. We use a finite difference approximation of the slope of $f(\alpha)$ to find its maximum. The value of $\alpha$ can also be determined visually by plotting the objective function (Figure \ref{fig:densityerrorplot}D); the sharp inflection point in the plot represents the value of $\alpha$. Algorithm \ref{alg:pulscar} shows the pseudocode of the PULSCAR algorithm to estimate $\alpha$ using the objective function based on probability densities. Algorithm \ref{alg:bandwidthestimate} is a subroutine to compute the beta kernel bandwidth. Full source code for our algorithms is available at GitHub (\url{https://github.com/unmtransinfo/PULSNAR/}).

Our PULSCAR algorithm  (Algorithm \ref{alg:pulscar}) uses a histogram bin count heuristic to generate a histogram-derived density, then optimizes the beta distribution kernel bandwidth to best fit that density. The approach relies on three key elements: use of the beta kernel for density estimates, the histogram bin count, and the kernel bandwidth. The beta kernel has important properties, described below, for fitting distributions over the interval $[0\ldots 1]$. The histogram bin count parameter is set using standard heuristics for fitting a histogram to data, and defines an array of evenly spaced numbers over the interval $[0\ldots 1]$, where the beta kernel density estimates are computed to optimize fitting the histogram. The bandwidth parameter determines the width of the beta kernel, balancing the need for narrower kernels to precisely estimate the distribution with many data points against wider kernels to avoid overfitting when data are sparse. The following three subsections detail these parameters and explain how they are determined.

\subsubsection*{Beta kernel density estimation}
\label{KernelBandwidthestimation}
A beta kernel estimator is used to create a smooth density estimate of both the positive and unlabeled ML probabilities, generating distributions over [$0 \dots1$], free of the problematic boundary biases of kernels (e.g. Gaussian) whose range extends outside that interval, adopting the approach of \citep{pulsnar_38}. Another problem with (faster) Gaussian kernel density implementations is that they often use polynomial approximations that can generate negative values in regions of low support, dramatically distorting $\alpha$ estimates, which require non-negative probability distribution estimates. The beta PDF is as follows \citep{pulsnar_41a}:
\begin{equation}
h(x, a, b) = \frac{\Gamma(a+b) x^{a-1} (1-x)^{b-1}}{\Gamma(a) \Gamma(b)},
\end{equation}
for x $\in$ [0,1], where $\Gamma$ is the gamma function, $a=1+\frac{z}{bw}$ and $b=1+\frac{1-z}{bw}$, with $z$ the bin edge (a value in an array of evenly spaced numbers over the interval [0, 1] with a total of $k$ bins), and $bw$ the bandwidth.

\subsubsection*{Histogram bin count} Our implementation supports 5 well-known methods to determine the number of histogram bins: square root, Sturges' rule, Rice's rule, Scott's rule, and Freedman–Diaconis (FD) rule \citep{pulsnar_35}. Let `pr' be the ML-predicted probabilities of PU examples and `n' be the number of  PU examples. The formulae to count the number of bins using these methods are as follows:
\begin{enumerate}[label=\roman*.]
  \item \textbf{Square Root Method: } $\text{Number of bins} = \sqrt{n}$
  \item \textbf{Sturges' Rule: } $\text{Number of bins} = 1 + \log_2(n)$
  \item \textbf{Rice's Rule: } $\text{Number of bins} = 2 \times n^{1/3}$
  \item \textbf{Scott's Rule: } \\
$\text{h} = 3.5 \times \frac{\text{StandardDeviation}(pr)}{n^{1/3}}$, 
$\text{Number of bins} = \frac{\text{max(pr)} - \text{min(pr)}}{\text{h}}$
\item \textbf{5. Freedman–Diaconis (FD) Rule: } \\
$\text{h} = 2 \times \frac{\text{InterQuartileRange(pr)}}{n^{1/3}}$, 
$\text{Number of bins} = \frac{\text{max(pr)} - \text{min(pr)}}{\text{h}}$
\end{enumerate}

\begin{minipage}[t]{.50\linewidth}
\begin{algorithm}[H]
    \caption{PULSCAR Algorithm}
    \label{alg:pulscar}
    \textbf{Input}: X ($X_p \cup X_u$), y ($y_p \cup y_u$), n\_bins \\
    \textbf{Output}: estimated $\alpha$
    \begin{algorithmic}[1] 
    \STATE predicted\_probabilities (p) $\leftarrow$ $\mathcal{A}(X,y)$
    \STATE p0 $\leftarrow$ p[y == 0]
    \STATE p1 $\leftarrow$ p[y == 1]
    \STATE estimation\_range $\leftarrow$ [0, 0.0001, 0.0002, ..., 1.0]
    \STATE bw $\leftarrow$ estimate\_bandwidth\_pu(p, n\_bins)
    \STATE $D_u$ $\leftarrow$ beta\_kernel(p0, bw, n\_bins)
    \STATE $D_p$ $\leftarrow$ beta\_kernel(p1, bw, n\_bins)
    \STATE $\epsilon \leftarrow |\min(D_p)|$
    \IF{$\epsilon$ = 0}
        \STATE $\epsilon \leftarrow 10^{-10}$
    \ENDIF
    \STATE len $\leftarrow$ length(estimation\_range)
    \STATE selected\_range $\leftarrow$ estimation\_range[2:len]
    \STATE $\alpha \leftarrow$ estimation\_range
    \STATE f($\alpha$) $\leftarrow \log(|\min(D_u - \alpha  D_p)| + \epsilon$)
    \STATE d $\leftarrow$ f'($\alpha$)
    \STATE i $\leftarrow$ where the value of d changes maximally 
    \STATE \textbf{return} selected\_range[i]
    \end{algorithmic}
\end{algorithm}
\end{minipage}
\hfill
\begin{minipage}[t]{.45\linewidth}
\vspace{0pt}
\begin{algorithm}[H]
    \caption{estimate\_bandwidth\_pu}
    \label{alg:bandwidthestimate}
    \textbf{Input}: predicted\_probabilities, n\_bins \\
    \textbf{Output}: bandwidth
    \begin{algorithmic}[1] 
    \STATE preds $\leftarrow$ predicted\_probabilities
    \STATE bw $\in$ [0.01, 0.5]
    \STATE $D_{hist}$ $\leftarrow$ histogram(preds, n\_bins, 
    \STATEx density=True)
    \STATE $D_{beta}$ $\leftarrow$ beta\_kernel(preds, bw, n\-bins)
    \STATE \textbf{return} optimize(MeanSquaredError($D_{hist}$, 
    \STATEx $D_{beta}$))
    \end{algorithmic}
    \end{algorithm}
\end{minipage}

\subsubsection*{Beta kernel bandwidth estimation} The bandwidth of the kernel is the smoothing parameter in kernel density estimation that affects the shape of the estimated distribution curve. A broader bandwidth results in a smoother and more generalized curve, whereas a narrower bandwidth produces a more fine-grained curve. Kernel bandwidth selection can also significantly influence $\alpha$ estimates: too narrow of a bandwidth can result in outliers driving poor estimates, and too wide of a bandwidth prevents distinguishing between distributions. We compute a histogram-derived density using the aforementioned bin count heuristic and beta kernel density estimate at those bin centers using the ML probabilities of both the positive and unlabeled examples. We find the minimum of the mean squared error (MSE) between the histogram and beta kernel densities using the scipy \textit{differential\_evolution()} optimizer \citep{pulsnar_41}, solving for the best bandwidth in the range [0.01...0.5]. The \textit{differential\_evolution()} algorithm is a stochastic direct search method to find the minimum of a multivariate function without requiring gradient information. The estimated bandwidth is chosen for kernel density estimation in the PULSCAR algorithm. All experiments herein use MSE as the error metric, but alternatively, the Jensen-Shannon distance can be employed.

\subsection*{Positive and Unlabeled Learning Selected Not At Random (PULSNAR) Algorithm}
We propose a new PU learning algorithm (PULSNAR) to estimate the  $\alpha$ in SNAR data, i.e., labeled positives are not selected completely at random. PULSNAR uses a divide-and-conquer strategy for the SNAR data to convert a SNAR problem into several sub-problems. It employs clustering techniques to divide SNAR positives into clusters, where each cluster predominantly contains one subtype of positives. Subsequently, it applies the PULSCAR algorithm to the positives from each cluster along with all unlabeled data to estimate the $\alpha$ for each subtype. The final $\alpha$ is computed by summing the $\alpha$ returned by the PULSCAR algorithm for each subtype of positives.
\begin{gather}
 \alpha = \alpha_1 + \alpha_2 + ... + \alpha_c,  \quad \text{$c=$ number of clusters}
\end{gather}

Figure \ref{fig:PULSNAR_flowchart} visualizes the PULSNAR algorithm, and Algorithm \ref{alg:pulsnar} provides its pseudocode.

\subsubsection*{Clustering rationale} 
Suppose both the positive and unlabeled sets contain positives from $k$ subclasses ($1\dots k$). With selection bias (SNAR), the subclass proportions will vary between the sets, and thus, the PDF of the labeled positives cannot be scaled by a uniform $\alpha$ to estimate positives in the unlabeled set. The smallest subclass would drive an $\alpha$ underestimate with PULSCAR. To address this, we perform clustering on the labeled positives using the key features that differentiated positives from the unlabeled. This assumes that these features also correlate with the factors driving selection bias within each subclass. By dividing positives into $c$ clusters based on these features, we establish that within each cluster, the selection bias is uniformly distributed, implying that all members of a cluster are equally likely to appear in the unlabeled set. This uniformity allows for the inference of a cluster-specific $\alpha$, which can be summed over all clusters to provide an overall estimate of positives in the unlabeled set (Figure \ref{fig:PULSNAR_flowchart}).

\subsubsection*{Determining the number of clusters in the positive set}
Determining the ``optimal number of clusters'' is a challenging problem, as evidenced by the NP-hardness of optimal k-means clustering in the literature \citep{pulsnar_54}. Our approach to determining the number of clusters via BIC is one of the widely used heuristics and approximation methods \citep{pulsnar_36, pulsnar_55, pulsnar_56}. We build an XGBoost \citep{pulsnar_5} model on all positive and unlabeled examples to determine the important features and their \emph{gain} scores. A \emph{gain} score measures the magnitude of the feature's contribution to the model. We select all labeled positives and then cluster them on those features scaled by their corresponding \emph{gain} score, using scikit\_learn's Gaussian mixture model (GMM) method. To establish the number of clusters (n\_components), we iterate n\_components over $1\ldots m$ (e.g., $m$=25) and compute the Bayesian information criterion (BIC)\citep{pulsnar_6} for each clustering model. We use max\_iter=250, and covariance\_type=``full''. The other parameters are used with their default values. We implemented the ``Knee Point Detection in BIC'' algorithm, explained in \citep{pulsnar_36}, to find the number of clusters in the labeled positives.

\subsection*{Calculating calibrated probabilities}
The approach to calibrate the ML-predicted probabilities of positive and unlabeled examples in the SCAR and SNAR data is explained in Appendix \ref{appendix1}.

\subsection*{Improving classification performance}
Enhancing classification with PULSCAR/PULSNAR involves estimating $\alpha$ and using the resulting calibrated probabilities to re-label the top $\alpha |U|$ unlabeled examples as positives, while treating the remainder of the unlabeled examples as negatives. The original positives retain their labels. Standard ML classification techniques (e.g. XGBoost) are then employed on these updated labels, as detailed in Appendix \ref{appendix2}.

\begin{minipage}[t]{.45\linewidth}
\vspace{0pt}
\begin{algorithm}[H]
    \caption{PULSNAR Algorithm}
    \label{alg:pulsnar}
    \textbf{Input}: X ($X_p \cup X_u$), y ($y_p \cup y_u$), n\_bins \\
    \textbf{Output}: estimated $\alpha$
    \begin{algorithmic}[1] 
    \STATE feature\_importance ($v_1...v_k$), imp\_features ($x_1...x_k$) $\leftarrow$ $\mathcal{A}(X,y)$ 
    \STATE $x'_1...x'_k$ $\leftarrow$ $x_1 v_1...x_k v_k$
    \STATE $X'_p$ $\leftarrow$ $X_p$[$x'_1...x'_k$]
    \STATE clusters $s_1...s_c$ $\leftarrow$ GMM($X'_p$)
    \STATE $\alpha$ $\leftarrow$ 0
    \FOR{c in $s_1...s_c$}
        \STATE X' $\leftarrow$ $X_p[c] \cup X_u$
        \STATE y' $\leftarrow$ $y_p[c] \cup y_u$
        \STATE $\alpha$ $\leftarrow$ $\alpha$ + PULSCAR(X', y', n\_bins)
    \ENDFOR
    \STATE \textbf{return} $\alpha$
    \end{algorithmic}
\end{algorithm}
\end{minipage}
\hfill
\begin{minipage}[t]{.50\linewidth}
\vspace{0pt}
\includegraphics[width=0.99\columnwidth]{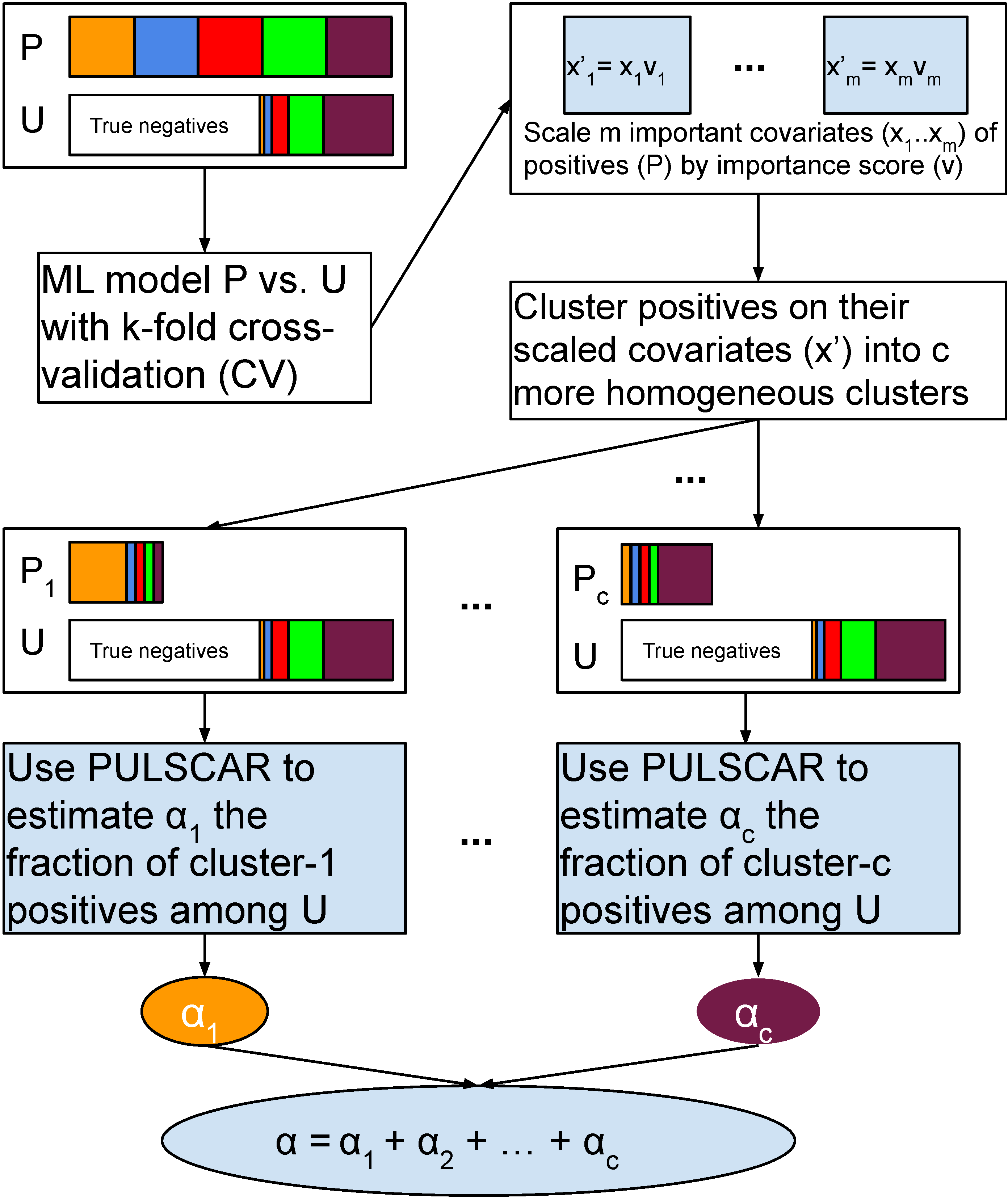}
\end{minipage}

\begin{figure}[H]
\caption{\textbf{Schematic of PULSNAR algorithm}. An ML model is trained and tested with 5-fold cross-validation (CV) on all positive and unlabeled examples. The important covariates that the model used are scaled by their importance value. Positives are divided into c clusters using the scaled important covariates. c ML models are trained and tested with 5-fold CV on the records from a cluster and all unlabeled records. We estimate the proportions ($\alpha_1...\alpha_c$) of each subtype of positives in the unlabeled examples using PULSCAR. The sum of those estimates gives the overall fraction of positives in the unlabeled set. P = positive set, U = Unlabeled set.}
\label{fig:PULSNAR_flowchart}
\end{figure}

\section*{Experimental Methods}
\label{sectionExperiments}
We evaluated our proposed PU learning algorithms in terms of $\alpha$ estimates, probability calibration (Appendix \ref{appendix1results}), and six classification performance metrics (Appendix \ref{appendix2results}). We used real-world ML benchmark datasets and synthetic data for our experiments. For real-world data, we used KDD Cup 2004 particle physics \citep{pulsnar_8}, Diabetes health indicators \citep{pulsnar_10a}, Magic gamma telescope \citep{pulsnar_10b}, Electrical grid stability \citep{pulsnar_10c}, Wilt \citep{pulsnar_10d}, and Mice protein expression \citep{pulsnar_10e} as SCAR datasets and Anuran calls \citep{pulsnar_10f}, Dry bean \citep{pulsnar_10g}, Room occupancy estimation \citep{pulsnar_10h}, Smartphone \citep{pulsnar_10i}, Letter recognition \citep{pulsnar_10j}, and
Statlog (Shuttle) \citep{pulsnar_9} as SNAR datasets. Synthetic (SCAR and SNAR) datasets were generated using the scikit-learn function \emph{make\_classification()} \citep{pulsnar_11}. We used XGBoost as a binary classifier in our proposed algorithms. We evaluated PU algorithms on imbalanced SCAR and SNAR datasets. In some datasets, we kept the number of unlabeled examples greater than the positive examples while in others, we kept it lesser. By setting the number of unlabeled examples to be less than positive examples, we were able to test the PU algorithms for a higher proportion of positives among unlabeled examples in ML benchmark datasets. To address the class imbalance, we used the \emph{scale\_pos\_weight} parameter of XGBoost to scale the weight of the labeled positive examples by the factor $s=\frac{|U|}{|P|}$. We also compared our methods with recently published methods for PU learning: KM1 and KM2 \citep{pulsnar_22}, TICE \citep{pulsnar_23} and DEDPUL \citep{pulsnar_24}. Some recent studies \citep{pulsnar_42, pulsnar_43} investigating scenarios where SCAR does not hold do not focus on $\alpha$ estimation or probability calibration. Therefore, we opted to exclude them from our comparison. KM1, KM2, and TICE algorithms were not scalable on large datasets, so we used relatively smaller synthetic and ML benchmark datasets to compare our methods with these methods. We compared PULSNAR with only DEDPUL on large synthetic datasets (Appendix \ref{appendix3}). Also, \citep{pulsnar_24} previously demonstrated that DEDPUL outperformed KM and TICE algorithms on several ML benchmark and synthetic datasets.

\subsection*{Synthetic data}
We generated SCAR and SNAR PU datasets with different fractions of positives (1\%, 5\%, 10\%, 20\%, 30\%, 40\%, and 50\%) among the unlabeled examples to test the effectiveness of our proposed algorithms. For each fraction, we generated 40 datasets using sklearn's
\emph{make\_classification()} function with random seeds 0-39. The \emph{class\_sep} parameter of the function was used to specify the separability of data classes. Values nearer to 1.0 make the classification task easier; we used class\_sep=0.3 to create difficult classification problems. 

\subsubsection*{SCAR data}
The datasets contained 2,000 positives (class 1) and 6,000 unlabeled (class 0) examples with 50 continuous features. The unlabeled set comprised $k\%$ positive examples with labels flipped to 0 and $(100-k)\%$ negative examples. 

\subsubsection*{SNAR data}
We generated datasets with 6 labels (0-5), defining `0' as negative and 1-5 as positive subclasses. These datasets contained 2,000 positives (400 from each positive subclass) and 6,000 unlabeled examples with 50 continuous features. The unlabeled set comprised k\% positive examples with labels (1-5) flipped to 0 and (100-k)\% negative examples. The unlabeled positives were markedly SNAR, with the 5 subclasses comprising 1/31, 2/31, 4/31, 8/31, and 16/31 of the unlabeled positives. (e.g., in the unlabeled set with 20\% positives, negative: 4,800, label 1 positive: 39, label 2 positive: 77, label 3 positive: 155, label 4 positive: 310, label 5 positive: 619).

\begin{table}[H]
\centering
\resizebox{\textwidth}{!}{%
\begin{tabular}{lllll}
\hline
\multicolumn{1}{|l|}{\textbf{SCAR Datasets}} &
  \multicolumn{1}{l|}{\textbf{Record count}} &
  \multicolumn{1}{l|}{\textbf{Feature count}} &
  \multicolumn{1}{l|}{\textbf{Positive class}} &
  \multicolumn{1}{l|}{\textbf{Negative class}} \\ \hline
\multicolumn{1}{|l|}{KDD Cup 2004 Particle Physics} &
  \multicolumn{1}{l|}{50,000} &
  \multicolumn{1}{l|}{77} &
  \multicolumn{1}{l|}{1} &
  \multicolumn{1}{l|}{0} \\ \hline
\multicolumn{1}{|l|}{CDC Diabetes Health Indicators} &
  \multicolumn{1}{l|}{253,680} &
  \multicolumn{1}{l|}{21} &
  \multicolumn{1}{l|}{0} &
  \multicolumn{1}{l|}{1} \\ \hline
\multicolumn{1}{|l|}{Magic Gamma Telescope} &
  \multicolumn{1}{l|}{19,020} &
  \multicolumn{1}{l|}{10} &
  \multicolumn{1}{l|}{g} &
  \multicolumn{1}{l|}{h} \\ \hline
\multicolumn{1}{|l|}{\begin{tabular}[c]{@{}l@{}}Electrical Grid Stability \\ Simulated Data\end{tabular}} &
  \multicolumn{1}{l|}{10,000} &
  \multicolumn{1}{l|}{13} &
  \multicolumn{1}{l|}{unstable} &
  \multicolumn{1}{l|}{stable} \\ \hline
\multicolumn{1}{|l|}{Wilt} &
  \multicolumn{1}{l|}{4,839} &
  \multicolumn{1}{l|}{5} &
  \multicolumn{1}{l|}{n} &
  \multicolumn{1}{l|}{w} \\ \hline
\multicolumn{1}{|l|}{Mice Protein Expression} &
  \multicolumn{1}{l|}{1,080} &
  \multicolumn{1}{l|}{77} &
  \multicolumn{1}{l|}{c-CS-m} &
  \multicolumn{1}{l|}{t-SC-s} \\ \hline
 &
   &
   &
   &
   \\ \hline
\multicolumn{1}{|l|}{\textbf{SNAR Datasets}} &
  \multicolumn{1}{l|}{\textbf{Record count}} &
  \multicolumn{1}{l|}{\textbf{Feature count}} &
  \multicolumn{1}{l|}{\textbf{Positive class}} &
  \multicolumn{1}{l|}{\textbf{Negative class}} \\ \hline
\multicolumn{1}{|l|}{Anuran Calls} &
  \multicolumn{1}{l|}{7,127} &
  \multicolumn{1}{l|}{23} &
  \multicolumn{1}{l|}{\begin{tabular}[c]{@{}l@{}}Leptodactylidae, \\ Dendrobatidae\end{tabular}} &
  \multicolumn{1}{l|}{Hylidae} \\ \hline
\multicolumn{1}{|l|}{Dry Bean} &
  \multicolumn{1}{l|}{13,611} &
  \multicolumn{1}{l|}{16} &
  \multicolumn{1}{l|}{\begin{tabular}[c]{@{}l@{}}SIRA, SEKER, HOROZ, \\ CALI, BARBUNYA, \\ BOMBAY\end{tabular}} &
  \multicolumn{1}{l|}{DERMASON} \\ \hline
\multicolumn{1}{|l|}{Room Occupancy Estimation} &
  \multicolumn{1}{l|}{10,129} &
  \multicolumn{1}{l|}{16} &
  \multicolumn{1}{l|}{0, 1, 3} &
  \multicolumn{1}{l|}{2} \\ \hline
\multicolumn{1}{|l|}{\begin{tabular}[c]{@{}l@{}}Smartphone Dataset for \\ Human Activity Recognition\end{tabular}} &
  \multicolumn{1}{l|}{7,415} &
  \multicolumn{1}{l|}{561} &
  \multicolumn{1}{l|}{1, 2, 3, 4, 6} &
  \multicolumn{1}{l|}{5} \\ \hline
\multicolumn{1}{|l|}{Letter Recognition} &
  \multicolumn{1}{l|}{4,639} &
  \multicolumn{1}{l|}{16} &
  \multicolumn{1}{l|}{A, B, C, E, F} &
  \multicolumn{1}{l|}{D} \\ \hline
\multicolumn{1}{|l|}{Statlog (Shuttle)} &
  \multicolumn{1}{l|}{43,500} &
  \multicolumn{1}{l|}{9} &
  \multicolumn{1}{l|}{2, 3, 4, 5, 6, 7} &
  \multicolumn{1}{l|}{1} \\ \hline
\end{tabular}%
}
\caption{List of SCAR and SNAR datasets with number of records and features. The table also shows which class was used as negative and which class(es) was/were used as positive. Unlabeled records included all records from the negative class and fractions of records from the positive class(es).}
\label{tab:uci-scar-snar-datasets}
\end{table}

\subsection*{ML Benchmark Datasets}
Table \ref{tab:uci-scar-snar-datasets} presents a comprehensive overview of the benchmark datasets used for testing PU methods. To add k\% positive examples to the unlabeled set, the labels of $m$ randomly selected positive records were flipped from 1 to 0, where $m = \frac{k |U|}{100-k}$. In order to satisfy the SNAR assumption, positive examples were selected from every positive subclass for label flipping while ensuring a distinct ratio of positive examples in the positive and unlabeled sets in SNAR datasets.

\subsection*{Estimation of fraction of positives among unlabeled examples}
We applied the PULSCAR algorithm to both SCAR and SNAR data, and the PULSNAR algorithm only to SNAR data, to estimate $\alpha$.

\subsubsection*{Using the PULSCAR algorithm}
To find the 95\% confidence interval (CI) on estimation, we ran XGBoost with 5-fold CV for 40 random instances of each dataset generated (or selected from benchmark data) using 40 random seeds. Each iteration's class 1 predicted probabilities of positives and unlabeled were used to calculate the value of $\alpha$.

\subsubsection*{Using the PULSNAR algorithm}
The labeled positives were divided into \emph{c} clusters to obtain homogeneous subclasses of labeled positives. The XGBoost ML models were trained and tested with 5-fold CV on data from each cluster and all unlabeled records. For each cluster, $\alpha$ was estimated by applying the PULSCAR method to class 1 predicted probabilities of positives from the cluster and all unlabeled examples. The overall proportion was calculated by summing the estimated $\alpha$ for each cluster. To compute the 95\% CI on the estimation, PULSNAR was repeated 40 times on data generated/selected using 40 random seeds.

\section*{Results}
\label{sectionResults}
\subsection*{Synthetic datasets}
Figure \ref{fig:synthetic_scar_non_scar_data} shows the $\alpha$ estimated by PU learning algorithms for synthetic datasets. TICE overestimated $\alpha$ for all fractions in both SCAR and SNAR datasets. For SCAR datasets, only PULSCAR returned close estimates for all fractions; DEDPUL overestimated for 1\%; KM1 overestimated for $\geq$ 20\%; KM2 underestimated for 50\% and overestimated for 1\%. For SNAR datasets, only PULSNAR's estimates were close to the true $\alpha$; other algorithms either overestimated or underestimated $\alpha$ for some or all fractions. Figure \ref{fig:dedpulVSpulsnar} in Appendix \ref{appendix3} shows the $\alpha$ estimated by DEDPUL and PULSNAR on large SNAR datasets with different class imbalances. As the class imbalance increased, DEDPUL underestimated $\alpha$, especially for larger fractions. The estimated $\alpha$ by the PULSNAR method was close to the true fractions across all sample sizes and class imbalances.

\subsection*{ML Benchmark datasets}
\subsubsection*{SCAR data}
Figure \ref{fig:uci_scar_data} shows the $\alpha$ estimated by PU learning algorithms for SCAR benchmark datasets. The KDD cup and diabetes datasets were too large to execute KM1 and KM2. These methods produced overestimated $\alpha$ values for the magic gamma and electric grid datasets, and for the wilt and mouse protein datasets, $\alpha$ values were either overestimated or underestimated for some fractions. TICE overestimated $\alpha$ for the KDD cup and mouse protein datasets, while for other datasets, it produced either overestimated or underestimated $\alpha$ values for some fractions. PULSCAR and DEDPUL both provided estimates that were close to the true answers for all fractions and datasets, but DEDPUL overestimated $\alpha$ for 1\% for all datasets.

\subsubsection*{SNAR data}

Figure \ref{fig:uci_non_scar_data} shows the $\alpha$ estimated by PU learning algorithms for SNAR benchmark datasets. For all datasets, only PULSNAR provided close estimates for all fractions, while other algorithms (KM1, KM2, TICE, DEDPUL, and PULSCAR) either overestimated or underestimated $\alpha$ for some or all fractions. For fractions $\geq15\%$, KM1, KM2, TICE, DEDPUL, and PULSCAR usually underestimated $\alpha$ across all datasets. The $\alpha$ estimates of KM1 were inaccurate for all fractions except 1\%, across all datasets. KM2 and TICE showed inconsistent results, overestimating or underestimating even for smaller fractions ($<10\%$) in some datasets. DEDPUL overestimated alpha for 1\% in all datasets. By contrast, our PULSCAR method provided $\alpha$ estimates that were close to the true answers for smaller fractions ($<10\%$).

\begin{figure}[H]
\centering
\includegraphics[width=5.75in]{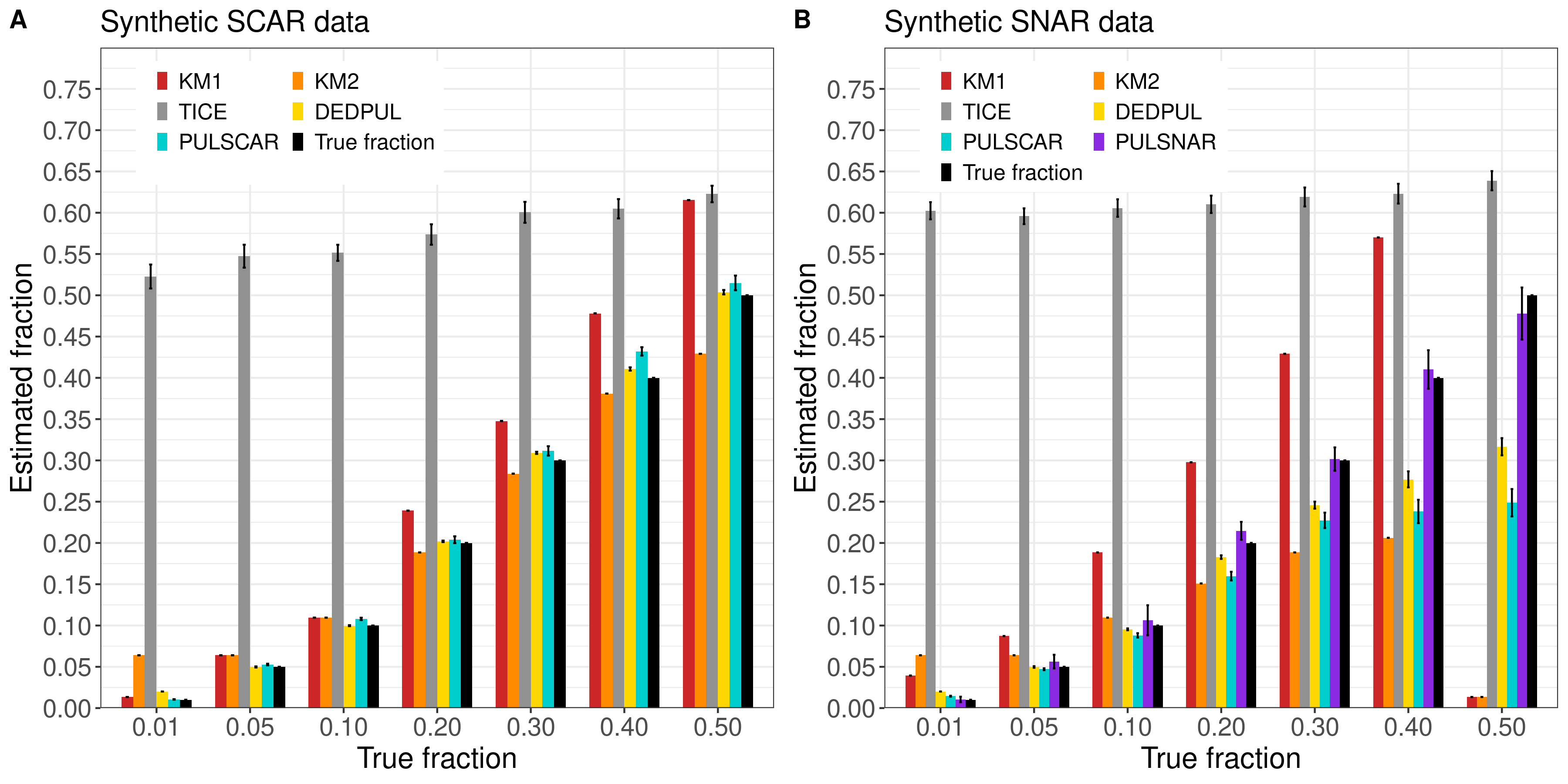} 
\caption{\textbf{KM1, KM2, TICE, DEDPUL, PULSCAR, and PULSNAR evaluated on SCAR and SNAR synthetic datasets}. The bar represents the mean value of the estimated $\alpha$, with 95\% confidence intervals for estimated $\alpha$. The best estimators are close to the black bars, representing the true $\alpha$. Bars larger than the black bars represent overestimation, while bars smaller than the black bars represent underestimation.}
\label{fig:synthetic_scar_non_scar_data}
\end{figure}

\subsection*{Execution time of PU methods}
Table \ref{tab:execution_time} presents the execution times of PU methods applied to synthetic SNAR and diabetes SCAR data. Despite PULSNAR determining the number of clusters and estimating $\alpha$ for each cluster, both PULSCAR and PULSNAR executed faster than other algorithms on large datasets. The specifications of the machine used to measure the execution time of these algorithms are as follows: Dual CPU AMD EPYC 9654 and 1.5TB RAM.

\begin{table}[h]
\centering
\resizebox{\textwidth}{!}{%
\begin{tabular}{|l|l|l|l|l|l|}
\hline
\textbf{Datasets}                                                                   & \textbf{KM} & \textbf{TICE} & \textbf{DEDPUL} & \textbf{PULSCAR} & \textbf{PULSNAR} \\ \hline
\begin{tabular}[c]{@{}l@{}}Synthetic SNAR data\\ (P: 5,000, U:100,000)\end{tabular} & Failed & 706.50              & 159.59      & 3.19        & 14.07  \\ \hline
\begin{tabular}[c]{@{}l@{}}Synthetic SNAR data\\ (P: 5,000, U:50,000)\end{tabular}  & Failed & 349.28  & 42.04        & 1.77        & 7.91  \\ \hline
\begin{tabular}[c]{@{}l@{}}Synthetic SNAR data\\ (P: 5,000, U:10,000)\end{tabular}  & 243.47   & 83.78      & 2.29      & 0.61        & 2.04    \\ \hline
\begin{tabular}[c]{@{}l@{}}Diabetes SCAR data\\ (P: 218,334, U: 35,346)\end{tabular} & Failed & 1100.73 & 199.88 & 18.31 & 27.85 \\ \hline
\end{tabular}%
}
\caption{\textbf{Execution time (in minutes) of PU algorithms on synthetic and benchmark datasets based on a single iteration to estimate $\alpha$ for all true fractions we tested.} For synthetic SNAR data, we tested these methods on true fractions [0.01, 0.05, 0.10, 0.20, 0.30, 0.40, 0.50], while for diabetes SCAR data, we tested these methods on true fractions [0.01, 0.05, 0.10, 0.20, 0.30, 0.40, 0.50, 0.60, 0.70, 0.80]. DEDPUL, PULSCAR, and PULSNAR were executed with 16 cores, whereas TICE and KM do not provide an option to set the number of CPU cores and use all available cores. The publicly available source code for KM algorithms is not equipped to manage large datasets, resulting in program crashes.}
\label{tab:execution_time}
\end{table}

\subsection*{Probability calibration}
Appendix \ref{appendix1results} contrasts the calibration curves for uncalibrated (blue) vs. isotonically calibrated (red) probabilities. The analysis further distinguishes whether the calibration included both original positives and unabeled examples, or just the unlabeled examples. Calibration for SCAR data utilized PULSCAR, while SNAR data calibration used PULSNAR. Calibration curves more closely aligned with the ideal $y=x$ line for SCAR data as compared to SNAR data. Moreover, calibration results were superior for larger fractions as opposed to smaller fractions (1\%) for both SCAR and SNAR data.

\begin{figure}[H]
\centering
\includegraphics[width=5.0in] {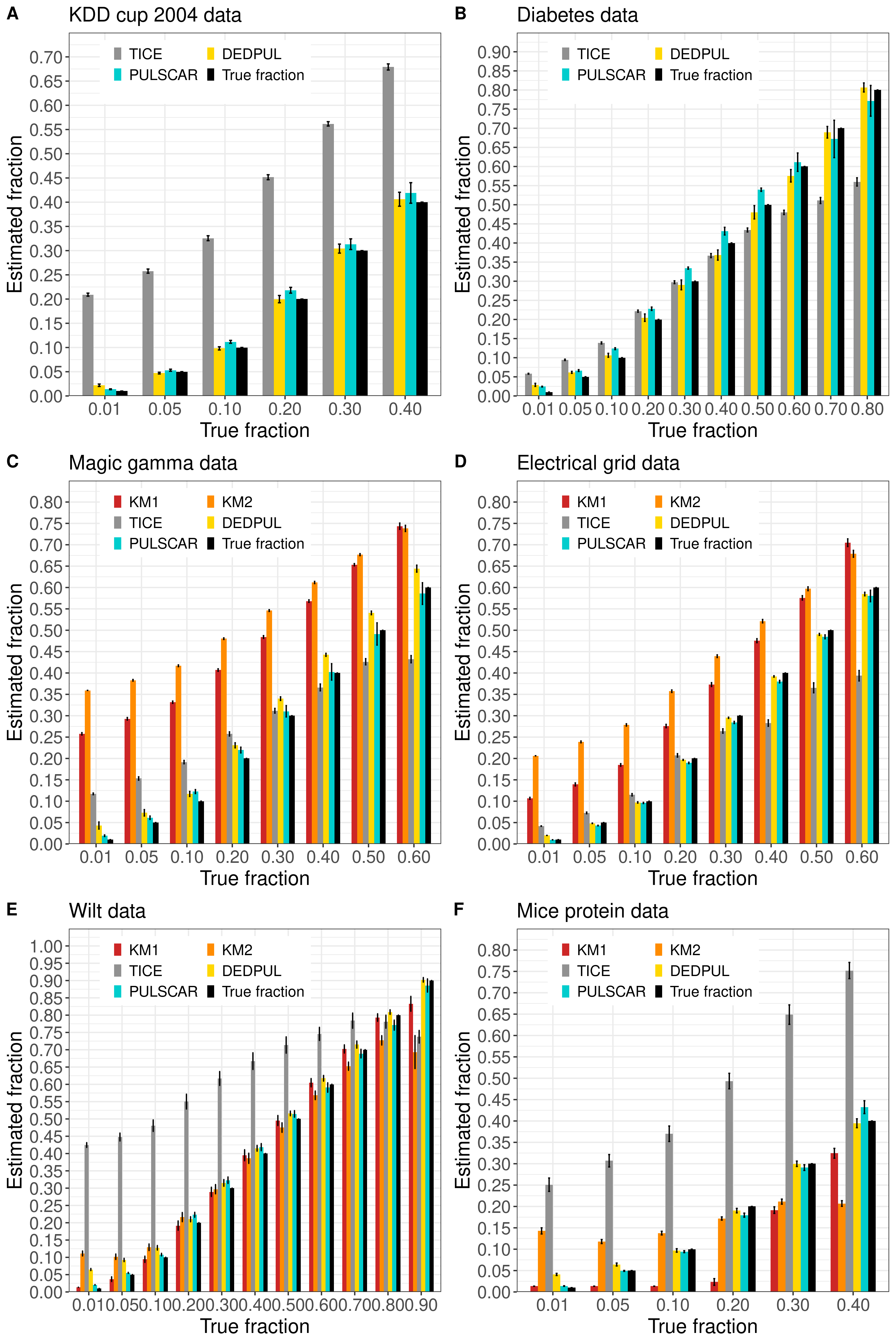}
\caption{\textbf{KM1, KM2, TICE, DEDPUL and PULSCAR evaluated on SCAR ML benchmark datasets}. The bar represents the mean value of the estimated $\alpha$, with 95\% confidence intervals for estimated $\alpha$. KM1 and KM2 failed to execute on the KDD cup and Diabetes datasets. The best estimators are close to the black bars, representing the true $\alpha$. Bars larger than the black bars represent overestimation, while bars smaller than the black bars represent underestimation.}
\label{fig:uci_scar_data}
\end{figure}

\begin{figure}[H]
\centering
\includegraphics[width=5.0in] {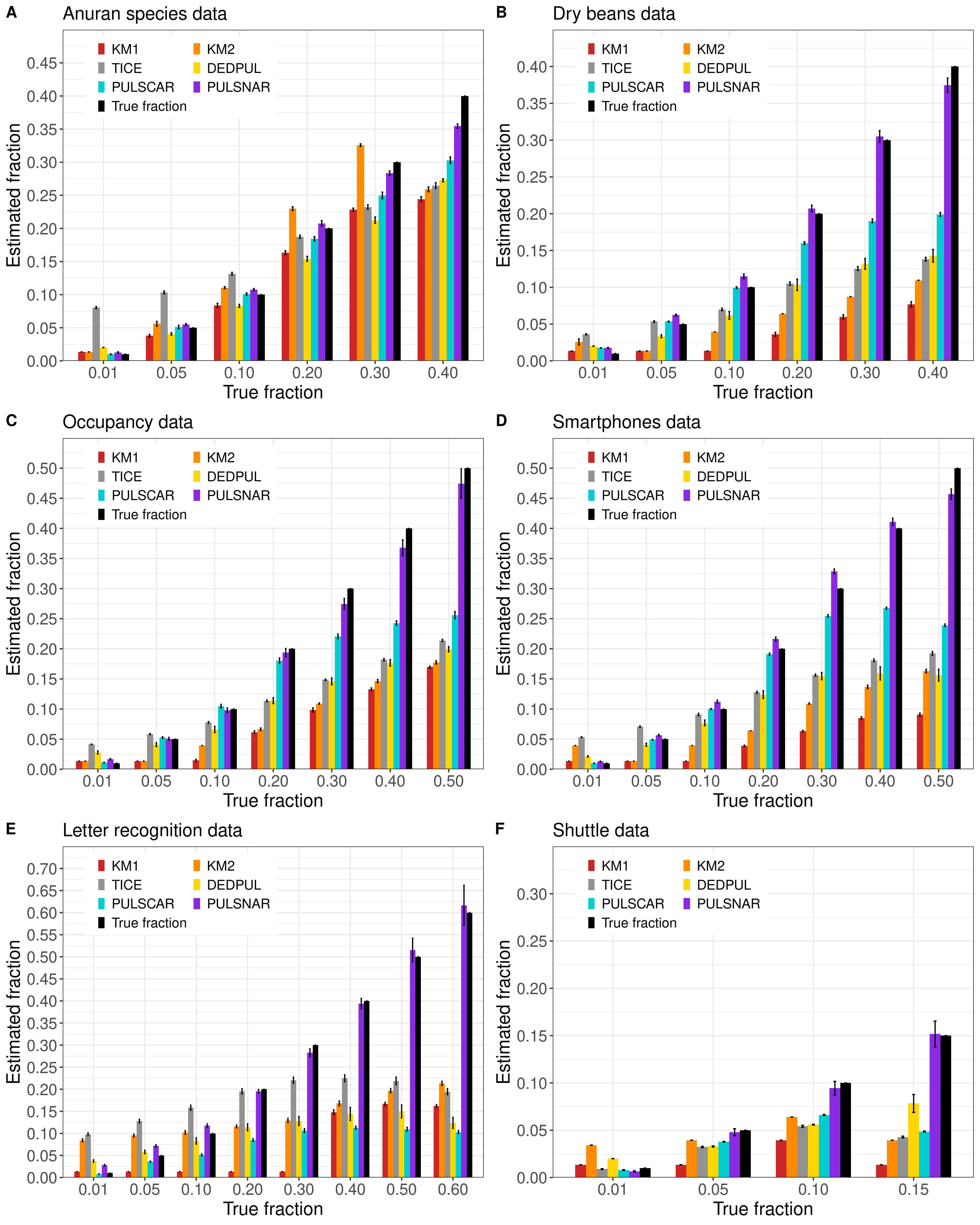}
\caption{\textbf{KM1, KM2, TICE, DEDPUL, PULSCAR and PULSNAR evaluated on SNAR ML benchmark datasets}. The bar represents the mean value of the estimated $\alpha$, with 95\% confidence intervals for estimated $\alpha$. As KM1 and KM2 were taking several hours to finish one iteration on the Shuttle dataset, the mean $\alpha$ was computed using 5 iterations, and the standard error was set to 0. The best estimators are close to the black bars, representing the true $\alpha$. Bars larger than the black bars represent overestimation, while bars smaller than the black bars represent underestimation.}
\label{fig:uci_non_scar_data}
\end{figure}

\subsection*{Classification performance metrics}
Appendix \ref{appendix2results} shows substantial improvement in 6 classification performance metrics when applying PULSCAR and PULSNAR versus XGBoost alone.  Appendix \ref{appendix4} presents the classification performance of PULSNAR and DEDPUL on two real-world SNAR datasets. The classification performance of DEDPUL declined for larger fractions due to the underestimation of $\alpha$. Conversely, PULSNAR exhibited robust classification performance due to its precise estimation of $\alpha$, which were close to true fractions.

\section*{Discussion and Conclusion}
\label{sectionConclusion}
This paper presented novel PU learning algorithms for estimating the proportion of positives ($\alpha$) among unlabeled examples in both SCAR and SNAR data with and without class imbalance. By utilizing the law of total probability to define the objective function (\ref{eq:obj_func}) and a beta kernel to estimate probability distributions, the PULSCAR algorithm estimates an upper bound on the fraction of positives in the unlabeled set under the assumption of SCAR and that the classes are separable. Although theoretically an upper bound, in practice, our estimates were all close to the actual ground truth $\alpha$, albeit with greater percent overestimation with smaller $\alpha$ values. We observed in all of our simulations on SCAR data that the $\alpha$ estimates were above ground truth, suggesting the PULSCAR algorithm is trustworthy for bounding $\alpha$ from above.

In our experiments, we demonstrated that our PULSCAR method outperformed other state-of-the-art methods, such as KM1, KM2, and TICE, for estimating $\alpha$ on both synthetic and real-world SCAR datasets (Figures \ref{fig:synthetic_scar_non_scar_data}A, \ref{fig:uci_scar_data}). The performance of the PULSCAR and DEDPUL methods was comparable on SCAR datasets, given their similar theoretical foundations. However, we observed that DEDPUL tends to have large overestimates for smaller fractions (e.g., for 1\%, DEDPUL estimated an $\alpha$ of $\geq2\%$ for all datasets). The version of PULSCAR presented in this paper uses the beta distribution, outperforming earlier prototypes that used the Gaussian distribution. This may account for PULSCAR's superiority over DEDPUL for 1\%, which uses the Gaussian kernel for density estimation. 

Our PULSNAR method demonstrated superior performance over all other methods (KM1, KM2, TICE, and DEDPUL) for estimating $\alpha$ across both synthetic and real-world SNAR datasets (Figures \ref{fig:synthetic_scar_non_scar_data}B, \ref{fig:uci_non_scar_data}, \ref{fig:dedpulVSpulsnar}). Additionally, the PULSNAR method showed better classification performance compared to DEDPUL when evaluated on SNAR datasets (Figures \ref{fig:dry_beans_classification}, \ref{fig:smartphones_classification}), particularly for larger fractions ($\geq20\%$). Interestingly, as the number of positive subclasses increased, the SCAR-based methods produced poor $\alpha$ estimates on SNAR data (Figures \ref{fig:synthetic_scar_non_scar_data}B, \ref{fig:uci_non_scar_data}), indicating that PU learning methods based on the SCAR assumption are not suitable for SNAR data. 

Probability calibration plays a crucial role in several classification tasks, such as medical diagnosis and financial risk assessment. It helps determine the optimal probability threshold for decision-making, which ultimately enhances the reliability and usefulness of ML models \citep{pulsnar_59, pulsnar_60}.
For instance, in a binary classification scenario where model predictions determine whether a patient has a disease, the selection of an optimal probability threshold for making informed decisions relies on having calibrated probabilities. Our study presented a novel approach to calculating calibrated probabilities for positive and unlabeled examples, which, to the best of our knowledge, has not been explored in existing PU methods. Our experimental results demonstrated that both PULSCAR and PULSNAR methods significantly improved probability calibration for SCAR and SNAR datasets, respectively (Appendix \ref{appendix1results}).

While not the main focus of this work, we also demonstrated that after applying PULSCAR and PULSNAR, classifier performance improved significantly (Appendix \ref{appendix2results}) for SCAR and SNAR data, respectively.

An important contribution of this work is the speed of $\alpha$ estimation. Our experimentation showed that the KM1, KM2, and TICE algorithms exhibited scalability issues. This observation aligns with the findings of \citep{pulsnar_51}, who noted the underperformance of these techniques in high-dimensional scenarios and scalability issues with large datasets. While we evaluated PULSCAR/PULSNAR against these methods using moderately sized datasets, it is plausible that their inherent limitations with data size and high dimension contributed to inaccurate $\alpha$ estimates for some of our test datasets. Considering the scalability limitations and long execution time associated with KM1, KM2, TICE, and DEDPUL (Table \ref{tab:execution_time}), these algorithms would be challenging to use for real-world applications involving millions of records and thousands of features. On the contrary, our preliminary work \citep{pulsnar_53} suggests that PULSCAR and PULSNAR are suitable for processing very large high-dimensional datasets ($n>1M$, 150,000+ features). While in principle, other SCAR-based algorithms can be substituted for PULSCAR into our PULSNAR clustering approach, due to overestimation/underestimation issues (Figure \ref{fig:synthetic_scar_non_scar_data}A, \ref{fig:uci_scar_data}), scalability concerns, maintenance issues, and execution time constraints (Table \ref{tab:execution_time}), employing KM1, KM2, TICE, and DEDPUL for estimating $\alpha$ in SNAR settings was not implemented in this work.

A limitation of our approach is that it relies on knowing whether the data are SCAR or SNAR to effectively select between PULSCAR or PULSNAR for $\alpha$ estimation. The issue is the ``knee point detection'' cluster determination approach relies on three consecutive points in the BIC curve to identify the knee (angle). As the first point cannot have a knee, this approach always returns more than one cluster. We tried other methods, such as minimum BIC, difference in BIC, or other clustering algorithms, but saw performance degradation in PULSNAR $\alpha$ estimates. When more than one clusters are used with PULSNAR on SCAR data, it can result in PULSNAR overestimating $\alpha$ as two near-identical positive types cannot be distinguished and get counted more than once. In the future, we aim to enhance the PULSNAR algorithm so that it can be applied without knowing whether the data are SCAR or SNAR. Importantly, preliminary work (not shown) indicates that PULSNAR $\alpha$ estimation is robust to overestimating the number of clusters in SNAR data: additional clusters beyond the known number of subtypes just slightly increase $\alpha$ estimates.

The utility of $\alpha$ estimation might be illustrated with the problem of estimating the fraction of bots among social media accounts, an active area of ML research \citep{pulsnar_57, pulsnar_58}. If a trustworthy $\alpha$ upper bound estimate for the bot fraction among unlabeled accounts came to, say, 20\%, business decisions could be made based on accounts comprising at least 80\% real people. However, one might expect that positively detected bots are not SCAR, and thus, PULSNAR provides a means for estimating $\alpha$ despite that. However, with the PULSNAR clustering heuristic, we cannot guarantee an upper bound as we can in the SCAR case.  Nevertheless, our experiments show that PULSNAR far outperformed SCAR-based methods with SNAR positives. In addition, we have found high utility with PULSNAR in identifying uncoded self-harm in electronic health records. In that setting, PULSCAR underestimated $\alpha$ due to self-harm positives being SNAR as expected, but with PULSNAR, as confirmed with chart review, we had accurate $\alpha$ estimation as well as good calibration \citep{pulsnar_53}.  Since the SCAR assumption frequently does not hold in real-world data, robust calibrated model predictions along with $\alpha$ estimates in SNAR settings using PULSNAR opens up new horizons in PU Learning.

\section*{Acknowledgments}
This research was supported by the National Institute of Mental Health of the National Institutes of Health under award numbers R01MH129764 and R56MH120826.

\bibliography{PULSNAR_paper}

\section*{Appendix}
\appendix

\section{Proof: positives are not independent of their attributes under the SNAR Assumption}
\label{appendix0}

Under the SNAR assumption, the probability that a positive example is labeled is not independent of its attributes. Stated formally, the assumption is that
$p(s=1|x, y=1) \neq p(s=1|y=1)$ i.e. $p(s=1|x, y=1)$ is not a constant.

\textbf{Proof:}
\begin{equation} 
\begin{split}
p(s=1|x, y=1) & = p(y=1|(s=1|x))p(s=1|x) \nonumber \\
 & = p(y=1|(s=1|x)) \frac{p(x|s=1)p(s=1)}{p(x)} \nonumber 
 \text{ , using Bayes' rule} \\
 & = \frac{p(x|s=1)p(s=1)}{p(x)} \nonumber
 \text{ , since $p(y=1|(s=1|x)) = 1$ } \\
 & = \text{a function of $x$.}
\end{split}
\end{equation}

\section{Algorithm for calibrating probabilities}
\label{appendix1}
Algorithm \ref{alg:calibprobs} shows the complete pseudocode to calibrate the machine learning (ML) model predicted probabilities. Once $\alpha$ is known, we seek to transform the original class 1 probabilities so that their sum is equal to $\alpha |U|$ among the unlabeled or $|P| + \alpha |U|$ among positive and unlabeled, and that they are well-calibrated. Our approach is to probabilistically flip $\alpha |U|$ labels of unlabeled to positive (from 0 to 1) in such a way as to match the PDF of labeled positives across 100 equispaced bins over $[0\ldots1]$, then fit a logistic or isotonic regression model on those labels versus the probabilities to generate the transformed probabilities. To determine the number of unlabeled examples that need to be flipped in each bin, we compute the normalized histogram density, $D\_hist$, for the labeled positives with 100 bins and then multiply $\alpha |U|$ with $D\_hist$.

The unlabeled examples are also divided into 100 bins based on their predicted probabilities. Starting from the bin with the highest probability(p=1), we randomly select \emph{k} examples and flip their labels from 0 to 1, where \emph{k} is the number of unlabeled examples that need to be flipped in the bin. If the number of records ($n_1$) that need to be flipped in a bin is more than the number of records ($n_2$) present in the bin, the difference ($n_1-n_2$) is added to the number of records to be flipped in the next bin, resulting in $\alpha |U|$ flips.

After flipping the labels of $\alpha |U|$ unlabeled examples from 0 to 1, we fit an isotonic or sigmoid regression model on the ML-predicted class 1 probabilities with the updated labels to obtain calibrated probabilities.

The above calibration approach applies to both SCAR and SNAR data. For the SNAR data, the PULSNAR algorithm divides labeled positive examples into \emph{k} clusters and estimates the $\alpha$ for each cluster. For each cluster, the ML-predicted class 1 probabilities of the examples (positives from the cluster and all unlabeled examples or only unlabeled examples) are calibrated using the estimated $\alpha$ for the cluster. Since, for each cluster, PULSNAR uses all unlabeled examples, each unlabeled example has \emph{k} ML-predicted/calibrated probabilities. The final ML-predicted/calibrated probability of an unlabeled example is calculated using the following Equation \ref{eq:combined_probs}:

\begin{equation}\label{eq:combined_probs}
p = 1 - (1-p_1)(1-p_2)\dots(1-p_k)
\end{equation} 
where $p_k$ is the probability of an unlabeled example from cluster \emph{k}.

\begin{algorithm}
\caption{calibrate\_probabilities}
\label{alg:calibprobs}
\textbf{Input}: predicted\_probs, labels, n\_bins, calibration\_method, calibration\_data, $\alpha$ \\
\textbf{Output}: calibrated\_probs
\begin{algorithmic}[1] 
\STATE p0 $\leftarrow$ predicted\_probs[labels == 0]
\STATE p1 $\leftarrow$ predicted\_probs[labels == 1]
\STATE y0 $\leftarrow$ labels[labels == 0]
\STATE y1 $\leftarrow$ labels[labels == 1]
\STATE $D_{hist}$ $\leftarrow$ histogram(p1, n\_bins, density=True)
\STATE unlab\_pos\_count\_in\_bin $\leftarrow$ $\alpha$  $|p0|$ $D_{hist}$
\STATE p0\_bins $\leftarrow$ split unlabeled examples into n\_bins using p0
\FOR{k $\leftarrow$ [n\_bins $\dots$ 1]}
    \STATE $n_1$ $\leftarrow$ unlab\_pos\_count\_in\_bin[k]
    \STATE $n_2$ $\leftarrow$ p0\_bins[k]
    \IF{$n_1 > n_2$}
        \STATE $\hat{y0}$ $\leftarrow$ flip labels (y0) of $n_2$ examples from 0 to 1 in bin k
        \STATE unlab\_pos\_count\_in\_bin[k-1] $\leftarrow$ unlab\_pos\_count\_in\_bin[k-1] + ($n_1-n_2$)
    \ELSE 
        \STATE $\hat{y0}$ $\leftarrow$ flip labels (y0) of random $n_1$ examples from 0 to 1 in bin k
    \ENDIF
\ENDFOR
\IF{calibration\_data == `PU'}
    \STATE p, y $\leftarrow$ $p1 \cup p0$, $y1 \cup \hat{y0}$
\ELSIF{calibration\_data == `U'}
    \STATE p, y $\leftarrow$ p0, $\hat{y0}$
\ENDIF
\IF{calibration\_method is `sigmoid'}
    \STATE $\hat{p}$ $\leftarrow$ LogisticRegression(p, y)
\ELSIF{calibration\_method is `isotonic'}
    \STATE $\hat{p}$ $\leftarrow$ IsotonicRegression(p, y)
\ENDIF
\STATE \textbf{return} $\hat{p}$
\end{algorithmic}
\end{algorithm}

\clearpage

\subsection{Experiments and Results}
\label{appendix1results}
We used synthetic SCAR and SNAR datasets and KDD Cup SCAR dataset to test our calibration algorithm. 

\textbf{SCAR datasets: }After estimating the $\alpha$ using the PULSCAR algorithm, we applied Algorithm \ref{alg:calibprobs} to calibrate the ML-predicted probabilities. To calculate the calibrated probabilities for both positive and unlabeled (PU) examples, we applied isotonic regression to the ML-predicted class 1 probabilities of PU examples with labels of positives and updated labels of unlabeled (of which $\alpha|U|$ were flipped per Algorithm \ref{alg:calibprobs}). We applied isotonic regression to the unlabeled's predicted probabilities with their updated labels to calculate the calibrated probabilities only for the unlabeled.

\textbf{SNAR datasets: } Using the PULSNAR algorithm, the labeled positive examples were divided into \emph{k} clusters. For each cluster, after estimating the $\alpha$, Algorithm \ref{alg:calibprobs} was used to calibrate the ML-predicted probabilities. To calculate the calibrated probabilities for positives from a cluster and all unlabeled examples, we applied isotonic regression to their ML-predicted class 1 probabilities with labels of positives from the cluster and updated labels of unlabeled (of which $\alpha_j |U|$ were flipped for cluster $j=1\dots k$, see Algorithm \ref{alg:calibprobs}). We applied isotonic regression to the unlabeled's predicted probabilities with their updated labels to calculate the calibrated probabilities only for the unlabeled. Thus, each unlabeled example had \emph{k} calibrated probabilities. We computed the final calibrated probability for each unlabeled example using Formula \ref{eq:combined_probs}.

\begin{figure}
\centering
\includegraphics[width=0.90\columnwidth]{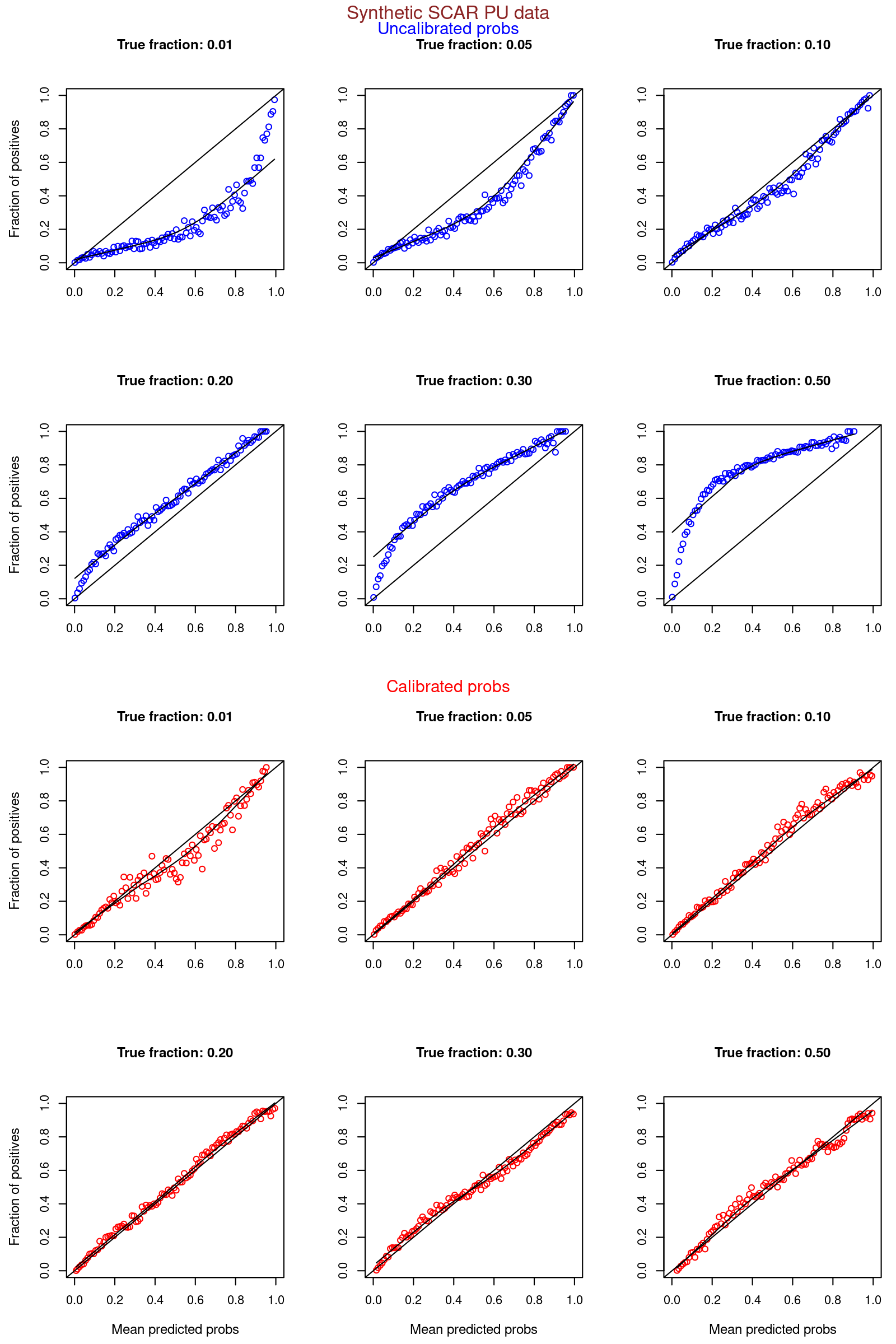}
\caption{\textbf{PULSCAR: calibration curves for synthetic SCAR datasets (both positive and unlabeled examples)}. Synthetic datasets were generated with different fractions of positives (1\%, 5\%, 10\%, 20\%, 30\%, and 50\%) among the unlabeled examples. class\_sep=0.3, number of attributes=100, $|P| = 5,000$ and $|U| = 50,000$. Calibration curves were generated using both positive and unlabeled examples (Uncalibrated probabilities - blue, calibrated probabilities - red).}
\label{fig:PU_scar_syn_pu_data_calibration}
\end{figure}

\begin{figure}
\centering
\includegraphics[width=0.90\columnwidth]{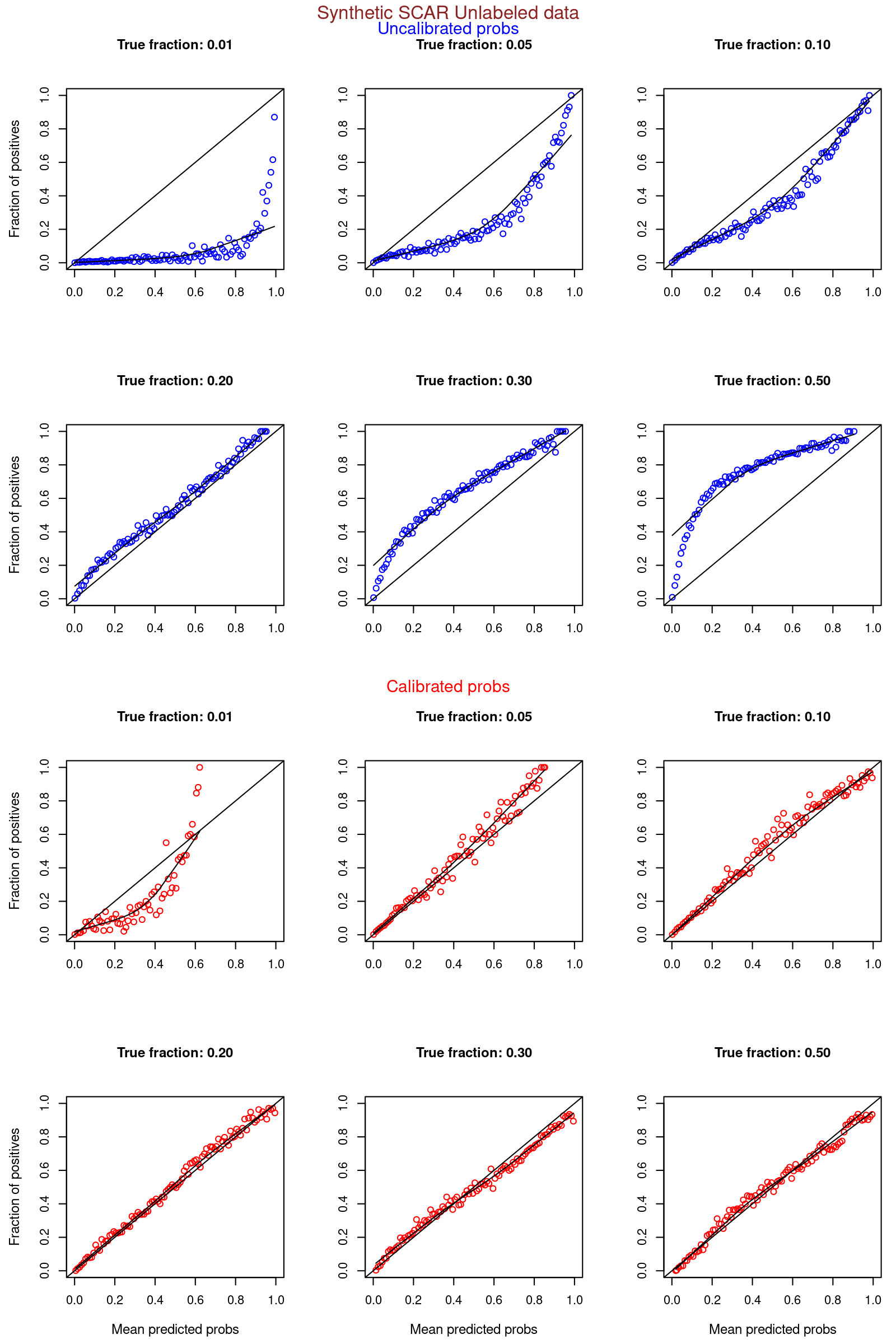}
\caption{\textbf{PULSCAR: calibration curves for synthetic SCAR datasets (only unlabeled examples)}. Synthetic datasets were generated with different fractions of positives (1\%, 5\%, 10\%, 20\%, 30\%, and 50\%) among the unlabeled examples. class\_sep=0.3, number of attributes=100, $|P| = 5,000$ and $|U| = 50,000$. Calibration curves were generated using only unlabeled examples (Uncalibrated probabilities - blue, calibrated probabilities - red).}
\label{fig:PU_scar_syn_u_data_calibration}
\end{figure}

\begin{figure}
\centering
\includegraphics[width=0.90\columnwidth]{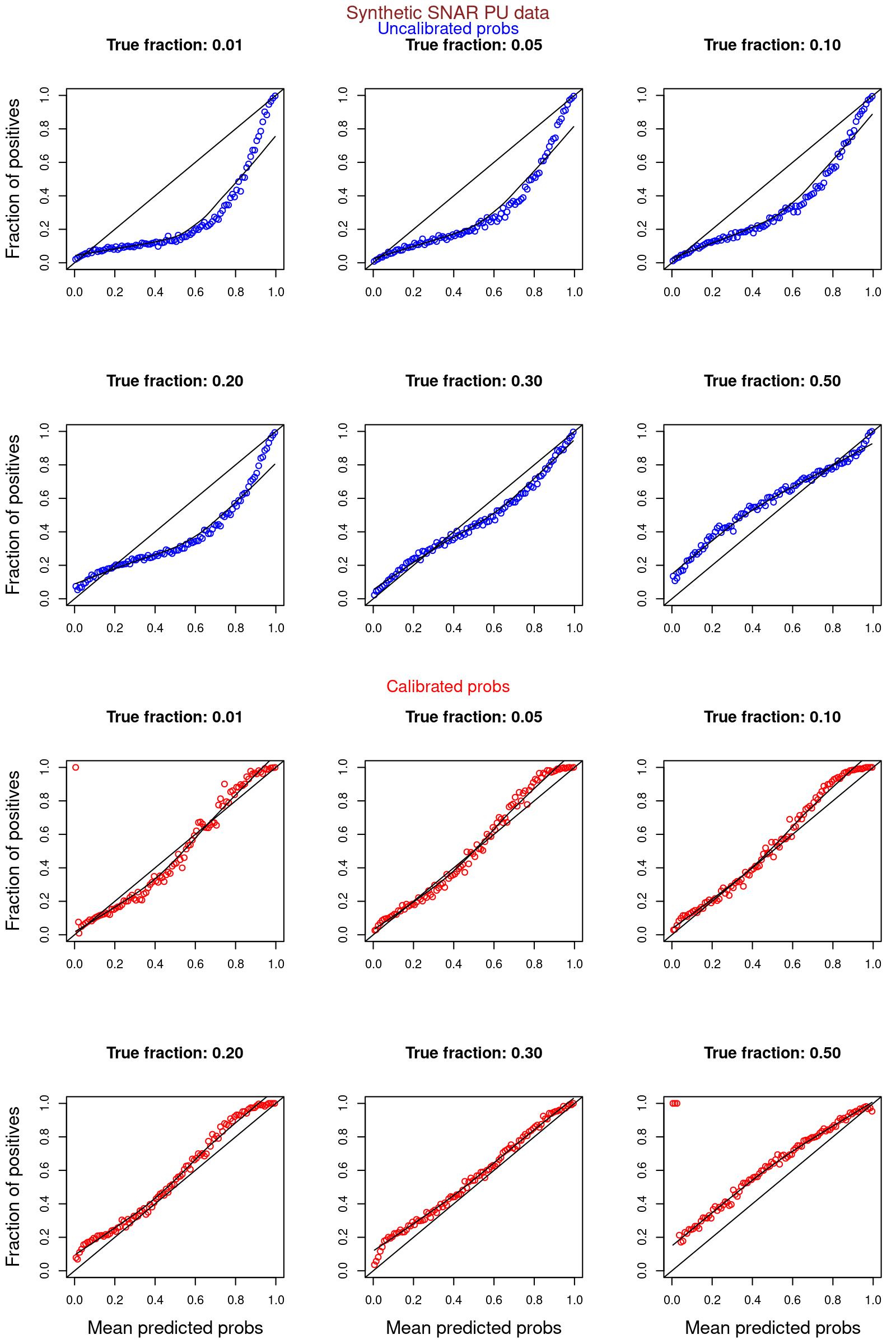}
\caption{\textbf{PULSNAR: calibration curves for synthetic SNAR datasets (both positive and unlabeled examples)}. Synthetic datasets were generated with different fractions of positives (1\%, 5\%, 10\%, 20\%, 30\%, and 50\%) among the unlabeled examples. class\_sep=0.3, number of attributes=100, number of positive subclasses=5, $|P|$ = 20,000 (4,000 from each subclass) and $|U|$ = 50,000. Calibration curves were generated using both positive and unlabeled examples (Uncalibrated probabilities - blue, calibrated probabilities - red).}
\label{fig:PU_snar_syn_pu_data_calibration}
\end{figure}

\begin{figure}
\centering
\includegraphics[width=0.90\columnwidth]{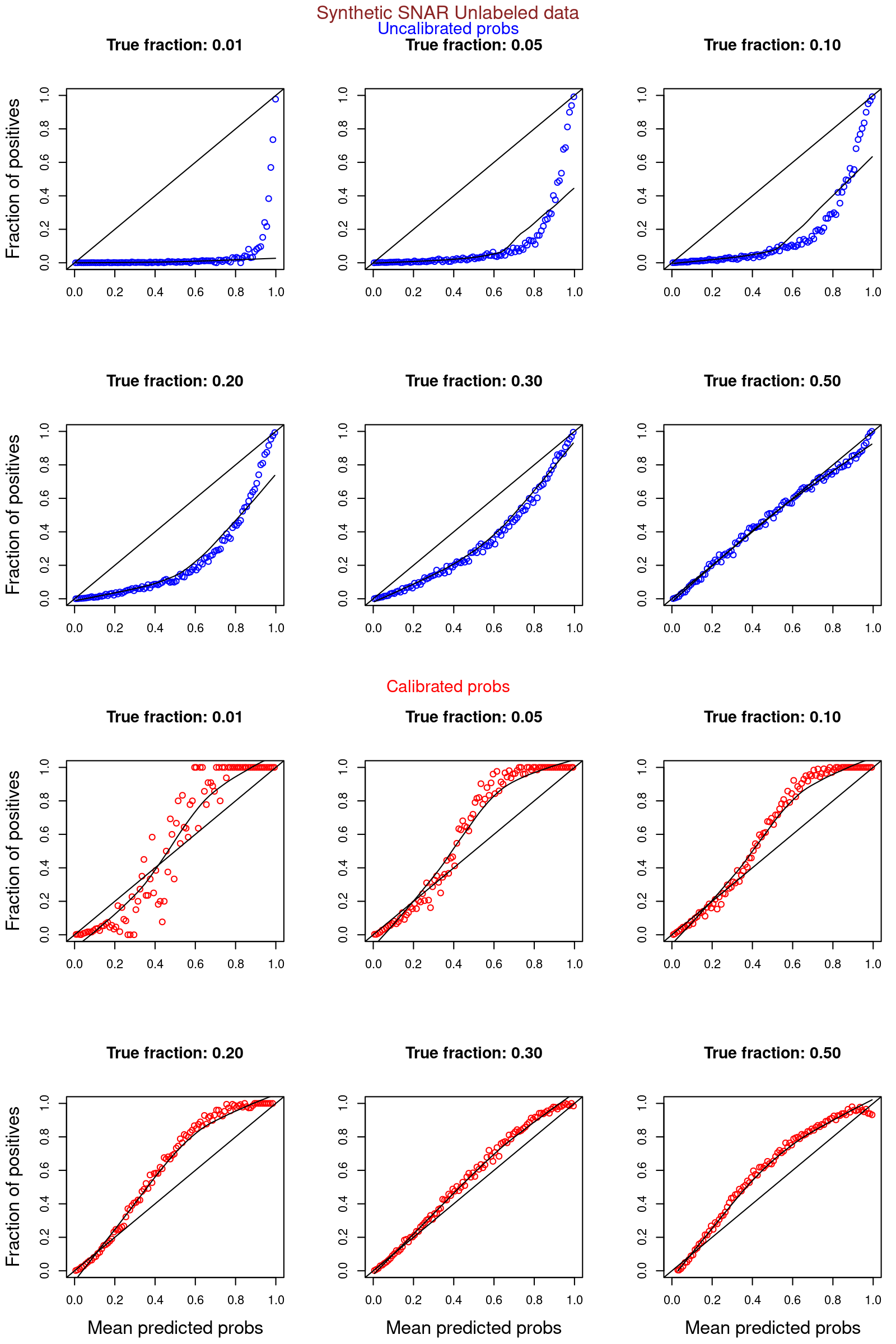}
\caption{\textbf{PULSNAR: calibration curves for synthetic SNAR datasets (only unlabeled examples)}. Synthetic datasets were generated with different fractions of positives (1\%, 5\%, 10\%, 20\%, 30\%, and 50\%) among the unlabeled examples. class\_sep=0.3, number of attributes=100, number of positive subclasses=5, $|P|$ = 20,000 (4,000 from each subclass) and $|U|$ = 50,000. Calibration curves were generated using only unlabeled examples (Uncalibrated probabilities - blue, calibrated probabilities - red).}
\label{fig:PU_snar_syn_u_data_calibration}
\end{figure}

\begin{figure}
\centering
\includegraphics[width=0.90\columnwidth]{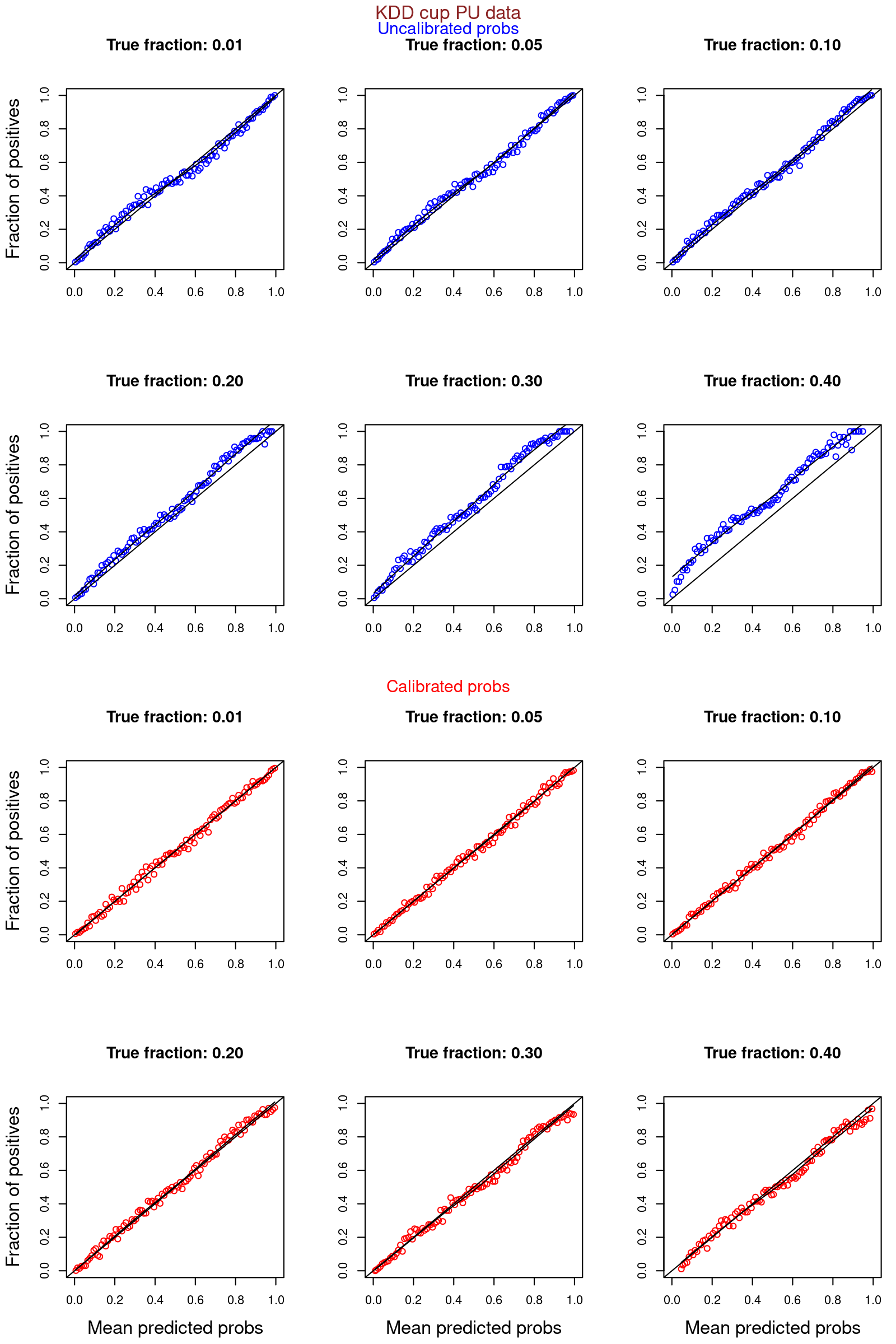}
\caption{\textbf{PULSCAR: calibration curves for SCAR KDD Cup 2004 particle physics dataset (both positive and unlabeled examples)}. Unlabeled sets contained 1\%, 5\%, 10\%, 20\%, 30\%, and 40\% positive examples. Calibration curves were generated using both positive and unlabeled examples (Uncalibrated probabilities - blue, calibrated probabilities - red).}
\label{fig:PU_scar_kdd_pu_data_calibration}
\end{figure}

\begin{figure}
\centering
\includegraphics[width=0.90\columnwidth]{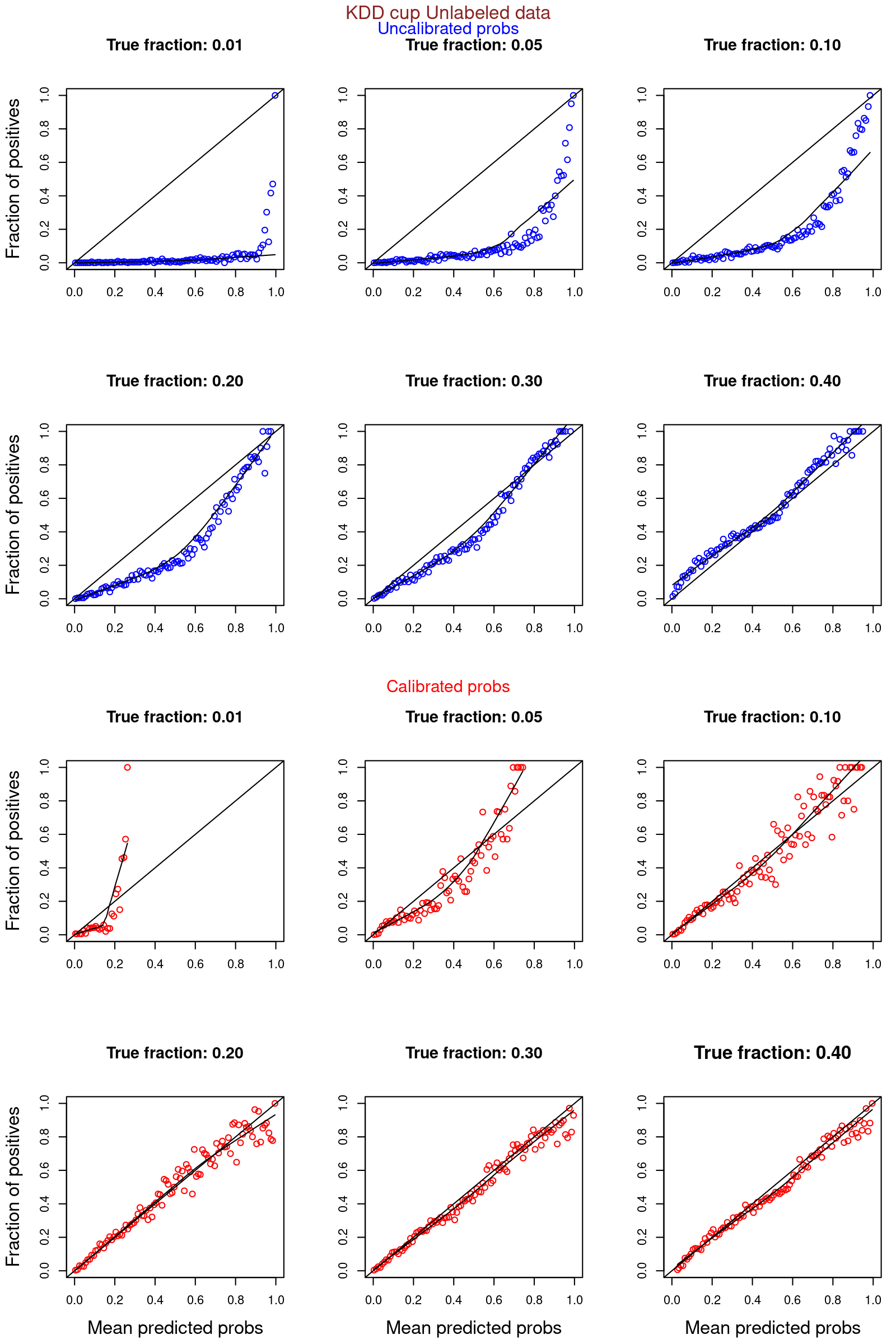}
\caption{\textbf{PULSCAR: calibration curves for SCAR KDD Cup 2004 particle physics dataset (only unlabeled examples)}. Unlabeled sets contained 1\%, 5\%, 10\%, 20\%, 30\%, and 40\% positive examples. Calibration curves were generated using only unlabeled examples (Uncalibrated probabilities - blue, calibrated probabilities - red).}
\label{fig:PU_scar_kdd_u_data_calibration}
\end{figure}

Figures \ref{fig:PU_scar_syn_pu_data_calibration}, \ref{fig:PU_scar_syn_u_data_calibration}, \ref{fig:PU_snar_syn_pu_data_calibration}, \ref{fig:PU_snar_syn_u_data_calibration}, \ref{fig:PU_scar_kdd_pu_data_calibration} and \ref{fig:PU_scar_kdd_u_data_calibration} show the calibration curves generated using the unblinded labels and isotonically calibrated (red)/ uncalibrated (blue) probabilities. When both positive and unlabeled examples were used to calculate calibrated probabilities, the calibration curve followed the y=x line (well-calibrated probabilities). When only unlabeled examples were used, the calibration curve for 1\% did not follow the y=x line, presumably due to the ML model being biased toward negatives, given the small $\alpha$. Also, the calibration curves for the SCAR data followed the y=x line more closely than the calibration curves for the SNAR data. It is due to the fact that the final probability of an unlabeled example in the SNAR data is computed using its \emph{k} probabilities from \emph{k} clusters. So, a poor probability estimate from even one cluster can influence the final probability of an unlabeled example. 

\clearpage

\clearpage
\section{Improving classification performance with PULSCAR and PULSNAR}
\label{appendix2}
Algorithm \ref{alg:classification_metrics} shows the complete pseudocode to improve classification performance with PULSCAR and PULSNAR. The algorithm returns the following six classification metrics: \emph{Accuracy, AUC-ROC, Brier score (BS), F1, Matthew's correlation coefficient (MCC)}, and \emph{Average precision score (APS)}. The approach to enhancing the classification performance is as follows: 

\textbf{Using PULSCAR: } After estimating the $\alpha$, the class 1 predicted probabilities of only unlabeled examples are calibrated using Algorithm \ref{alg:calibprobs}. The calibrated probabilities of the unlabeled examples are sorted in descending order, and the labels of top $\alpha |U|$ unlabeled examples with the highest calibrated probabilities are flipped from 0 to 1 (probable positives). We then train and test an ML classifier (XGBoost) with 5-fold CV using the labeled positives, probable positives, and the remaining unlabeled examples. The classification performance metrics are calculated using the ML predictions and the true labels of the data. 

\textbf{Using PULSNAR: } The PULSNAR algorithm divides the labeled positive examples into \emph{k} clusters. For each cluster, after estimating $\alpha_j$ for $j$ in $1\dots k$, the class 1 predicted probabilities of only unlabeled examples are calibrated using Algorithm \ref{alg:calibprobs}. Since each unlabeled example has \emph{k} calibrated probabilities, we compute the final calibrated probability for each unlabeled example using the Formula \ref{eq:combined_probs}. The final $\alpha$ is calculated by summing the $\alpha_j$ values over the $k$ clusters. The final calibrated probabilities of the unlabeled examples are sorted in descending order, and the labels of top $\alpha |U|$ unlabeled examples with the highest calibrated probabilities are flipped from 0 to 1 (probable positives). We then train and test an ML classifier (XGBoost) with 5-fold CV using the labeled positives, probable positives, and the remaining unlabeled examples. The classification performance metrics are calculated using the ML predictions and the true labels of the data.

\begin{algorithm}
    \caption{calculate\_classification\_metrics}
    \label{alg:classification_metrics}
    \textbf{Input}: X ($X_p \cup X_u$), y ($y_p \cup y_u$), y\_true, bin\_method, n\_bins, predicted\_probabilities, $\alpha$ \\
    \textbf{Output}: classification\_metrics (accuracy, roc auc, brier score, f1, Matthew's correlation coefficient, average precision) 
\begin{algorithmic}[1] 
    \STATE p $\leftarrow$ predicted\_probabilities
    \STATE $\hat{p}$ $\leftarrow$ calibrate\_probabilities(p, y, n\_bins, calibration\_method, `U', $\alpha$)
    \STATE sort $\hat{p}$ in descending order
    \STATE $\hat{y_u}$ $\leftarrow$ flip labels of top $\alpha |U|$ unlabeled examples with highest $\hat{p}$
    \STATE y $\leftarrow$ $y_p \cup \hat{y_u}$
    \STATE predicted\_probabilities (p) $\leftarrow$ $\mathcal{A}(X,y)$
    \STATE \textbf{return} accuracy(p, y\_true), auc(p, y\_true), bs(p, y\_true), f1(p, y\_true), mcc(p, y\_true), aps(p, y\_true)
\end{algorithmic}    
\end{algorithm}

\subsection{Experiments and Results}
\label{appendix2results}
We applied Algorithm \ref{alg:classification_metrics} to synthetic SCAR and SNAR datasets to get the performance metrics for the XGBoost model with PULSCAR and PULSNAR, respectively. The classification performance metrics were also calculated without applying the PULSCAR or PULSNAR algorithm, in order to determine the improvement in the classification performance of the model. The experiment was repeated 40 times by selecting different train and test sets using 40 random seeds to compute the 95\% confidence interval (CI) for the metrics. 

\begin{figure}
\centering
\includegraphics[width=0.90\columnwidth]{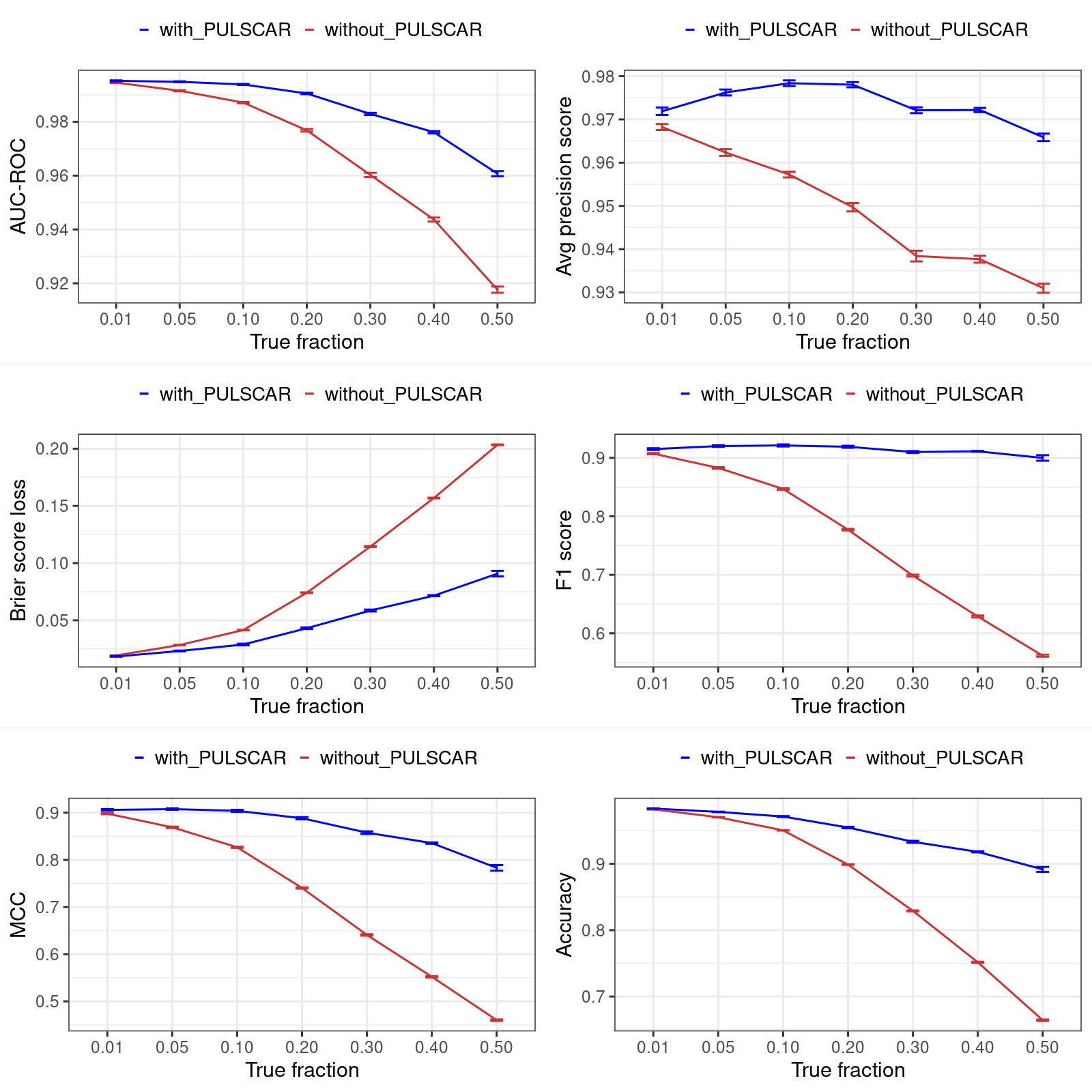}
\caption{\textbf{Classification performance of XGBoost model on synthetic SCAR datasets with and without the PULSCAR algorithm}. Synthetic datasets were generated with different fractions of positives (1\%, 5\%, 10\%, 20\%, 30\%, 40\%, and 50\%) among the unlabeled examples. class\_sep=0.3, number of attributes=100, $|P| = 5,000$ and $|U| = 50,000$. \emph{``without PULSCAR''} (brown): XGBoost model was trained and tested with 5-fold CV on the given data; the classification metrics were calculated using the model predictions and true labels. \emph{``with PULSCAR''} (blue): PULSCAR algorithm was used to find the proportion of positives among unlabeled examples ($\alpha$); using $\alpha$, probable positives were identified; XGBoost model was trained and tested with 5-fold CV on labeled positives, probable positives, and the remaining unlabeled examples; classification metrics were calculated using the model predictions and true labels. The error bars represent 95\% CIs for the performance metrics.}
\label{fig:scar_syn_classification_metrics}
\end{figure}

\begin{figure}
\centering
\includegraphics[width=0.90\columnwidth]{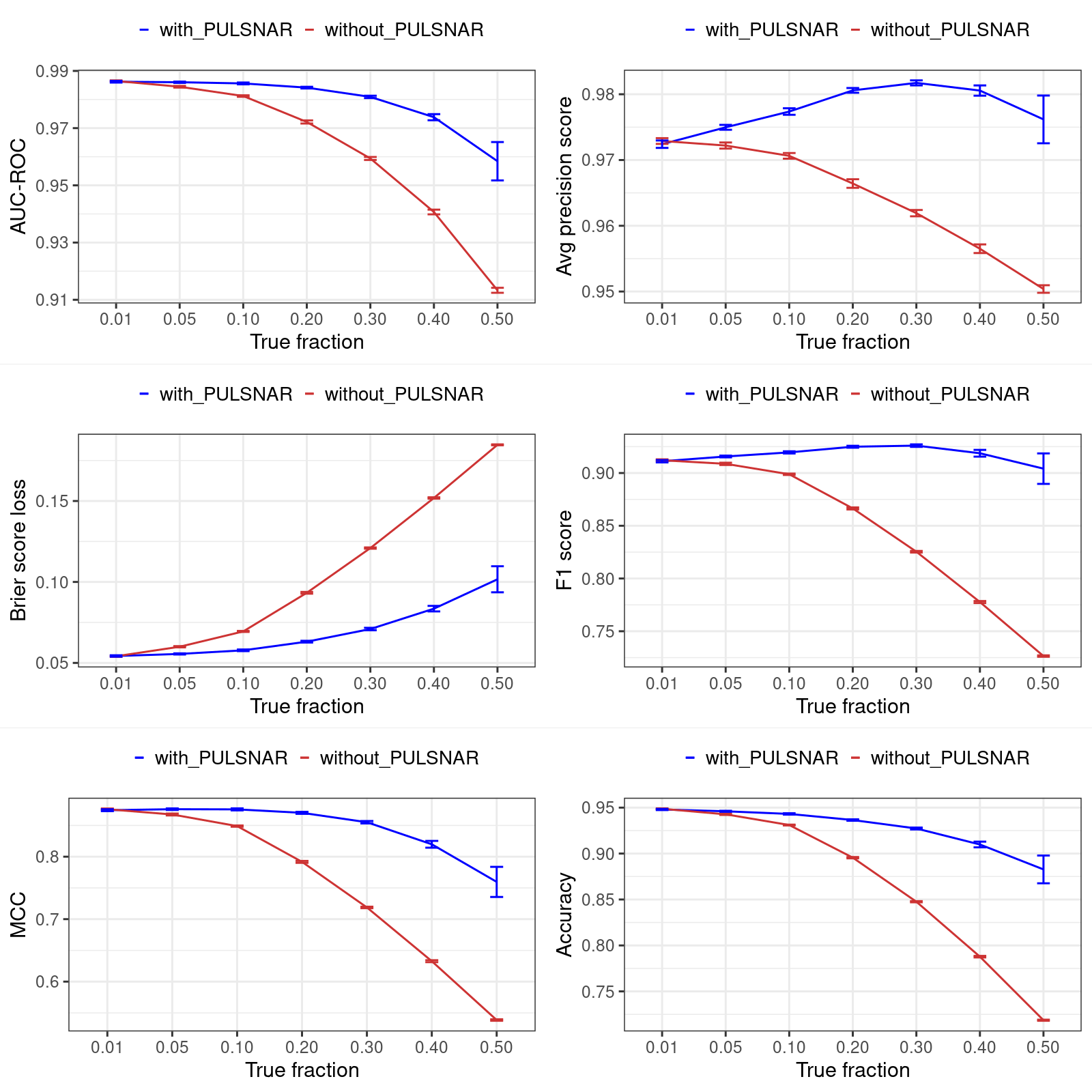}
\caption{\textbf{Classification performance of XGBoost model on synthetic SNAR datasets with and without the PULSNAR algorithm}. Synthetic datasets were generated with different fractions of positives (1\%, 5\%, 10\%, 20\%, 30\%, 40\%, and 50\%) among the unlabeled examples. class\_sep=0.3, number of attributes=100, number of positive subclasses=5, $|P|$ = 20,000 (4,000 from each subclass) and $|U|$ = 50,000. \emph{``without PULSNAR''} (brown): XGBoost model was trained and tested with 5-fold CV on the given data; the classification metrics were calculated using the model predictions and true labels. \emph{``with PULSNAR''} (blue): PULSNAR algorithm was used to find the proportion of positives among unlabeled examples ($\alpha$); using $\alpha$, probable positives were identified; XGBoost model was trained and tested with 5-fold CV on labeled positives, probable positives, and the remaining unlabeled examples; classification metrics were calculated using the model predictions and true labels. The error bars represent 95\% CIs for the performance metrics.}
\label{fig:snar_syn_classification_metrics}
\end{figure}

Figures \ref{fig:scar_syn_classification_metrics} and \ref{fig:snar_syn_classification_metrics} show the classification performance of the XGBoost model with/without the PULSCAR or PULSNAR algorithm on synthetic SCAR and SNAR data, respectively. The classification performance using PULSCAR or PULSNAR increased significantly over XGBoost alone. As the proportion of positives among the unlabeled examples increased, the performance of the model without PULSCAR or PULSNAR (blue) worsened significantly more than when using PULSCAR or PULSNAR.

\clearpage
\section{DEDPUL vs. PULSNAR: Alpha estimation}
\label{appendix3}
Public implementations of the PU learning methods KM1, KM2, and TICE were not scalable; they either failed to execute or would have taken weeks to run the multiple iterations required to obtain confidence estimates for large datasets. We thus could not compare our method with KM1, KM2, and TICE on large datasets and used only DEDPUL for comparison. Importantly, it was previously demonstrated that the DEDPUL method outperformed these three methods on several ML benchmark and synthetic datasets \citep{pulsnar_24}. 

We compared our algorithm with DEDPUL on synthetic SNAR datasets with different fractions (1\%, 5\%, 10\%, 20\%, 30\%, 40\%, and 50\%) of positives among unlabeled examples. In our experiments, we observed that class imbalance (ratio of majority class to minority class) could affect the $\alpha$ estimates. So, we used 4 different sample sizes: 1) positive: 5,000 and unlabeled: 5,000; 2) positive: 5,000 and unlabeled: 25,000; 3) positive: 5,000 and unlabeled: 50,000; 4) positive: 5,000 and unlabeled: 100,000. For each sample size and fraction, we generated 20 datasets using sklearn's \emph{make\_classification()} method with random seeds 0-19 to compute 95\% CI. We used class\_sep=0.3 for each dataset to create difficult classification problems. All datasets were generated with 100 attributes and 6 labels (0-5), defining `0' as negative and 1-5 as positive subclasses. The positive set contained 1000 examples from each positive subclass in all datasets. The unlabeled set comprised k\% positive examples with labels (1-5) flipped to 0 and (100-k)\% negative examples. The unlabeled positives were markedly SNAR, with the 5 subclasses comprising 1/31, 2/31, 4/31, 8/31, and 16/31 of the unlabeled positives. 

\begin{figure}
\centering
\includegraphics[width=0.95\columnwidth]{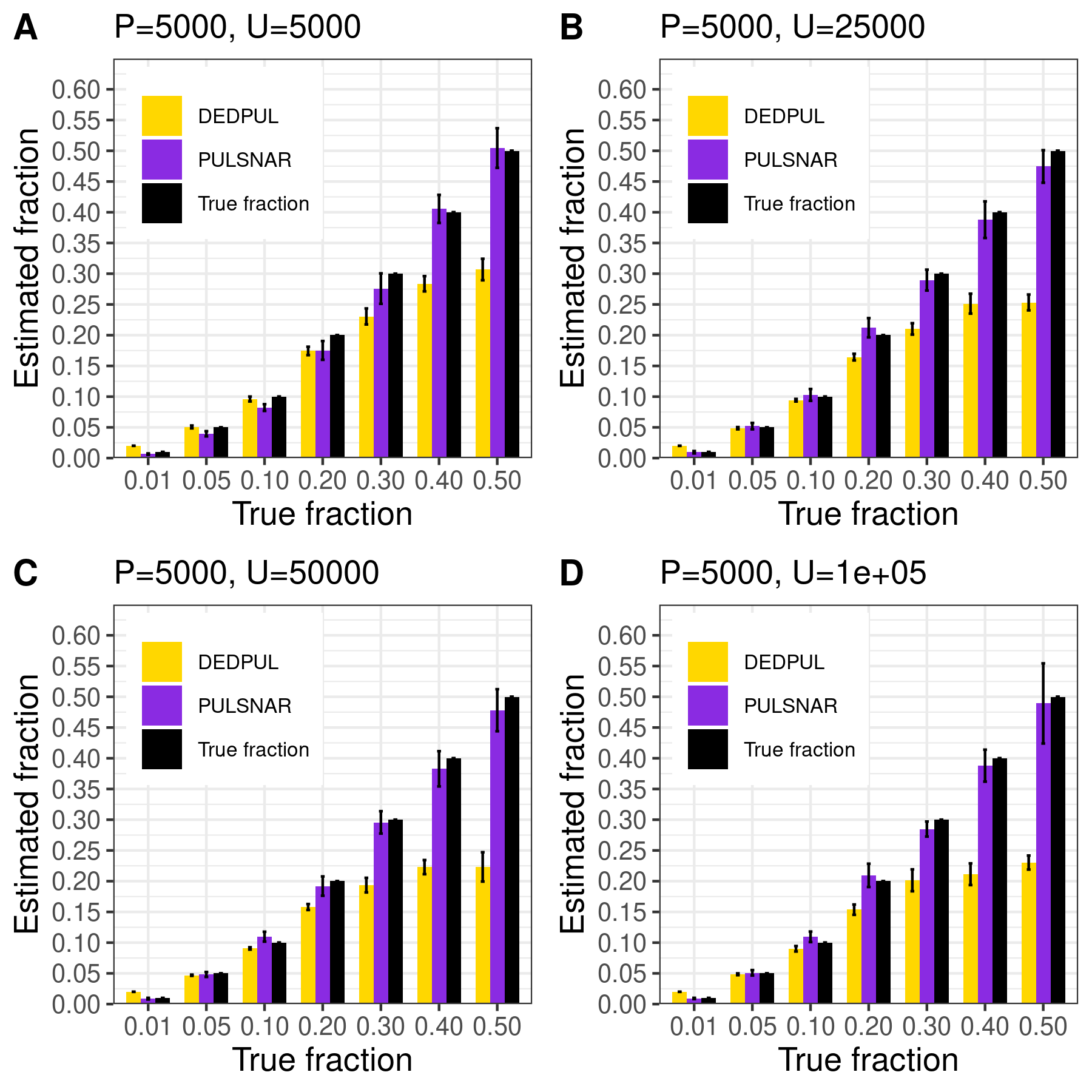}
\caption{\textbf{PULSNAR and DEDPUL evaluated on synthetic SNAR datasets}. The bar represents the mean value of the estimated $\alpha$, with 95\% CI for estimated $\alpha$. The best estimators are close to the black bars, representing the true $\alpha$. Bars larger than the black bars represent overestimation, while bars smaller than the black bars represent underestimation.}
\label{fig:dedpulVSpulsnar}
\end{figure}

Figure \ref{fig:dedpulVSpulsnar} shows the $\alpha$ estimates by DEDPUL and PULSNAR on synthetic SNAR data. For smaller true fractions (1\%, 5\%, 10\%), DEDPUL returned close $\alpha$ estimates, but for larger fractions (20\%, 30\%, 40\%, and 50\%), it underestimated $\alpha$. Also, as the class imbalance increased, the performance of DEDPUL dropped, especially for larger true fractions. The estimated $\alpha$ by the PULSNAR method was close to the true $\alpha$ for all fractions and sample sizes. 

\clearpage
\section{DEDPUL vs. PULSNAR: Classification performance}
\label{appendix4}
The main focus of this study was on comparing $\alpha$ estimates provided by different PU methods. Hence, we did not extensively compare the classification performance of our methods with other PU methods. Moreover, methods such as KM1, KM2, and TICE do not return predicted probabilities, which hinders the comparison of classification performance with these methods. Also, the classification performance of PULSCAR and DEDPUL on SCAR datasets was comparable. Consequently, we compared the classification performance of PULSNAR and DEDPUL using two SNAR datasets: smartphone and dry bean. Future studies would focus on a more comprehensive comparison of classification performance and exploring further enhancements in this regard. 

For DEDPUL, leveraging its estimated $\alpha$, we flipped the labels of $\alpha|U|$ unlabeled examples with the highest predicted probabilities from 0 to 1. Subsequently, we employed the XGBoost model with 5-fold CV, treating true positives and probable positives as positive examples, and probable negatives as negative examples to generate predicted probabilities for all examples in the datasets. The classification performance metrics were then computed using these XGBoost predicted probabilities and the true labels of the examples.

For PULSNAR, we employed the method discussed in Appendix \ref{appendix2} to calculate the classification performance metrics.

\begin{figure}
\centering
\includegraphics[width=0.95\columnwidth]{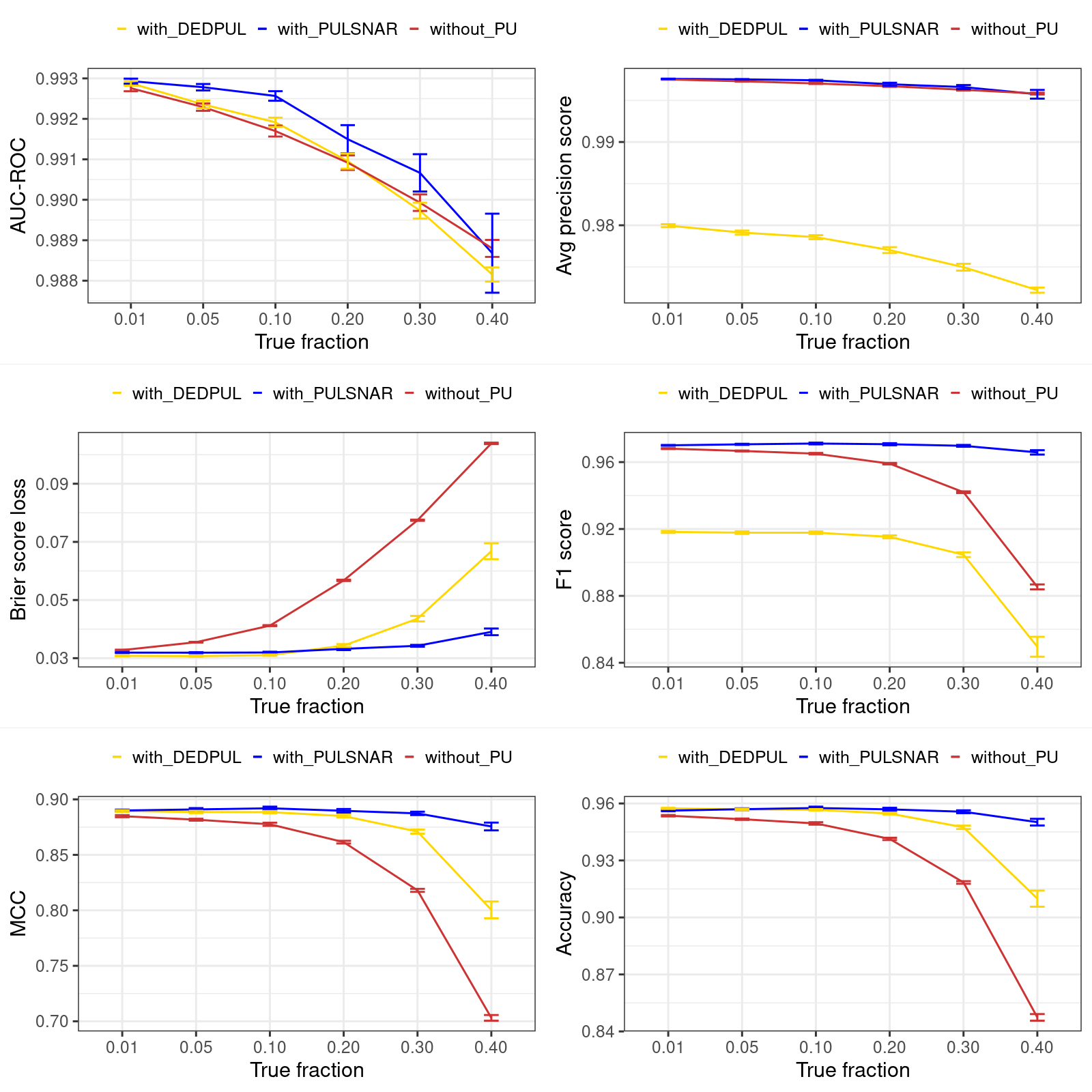}
\caption{\textbf{Classification performance of XGBoost model on dry bean SNAR dataset with/without DEDPUL and PULSNAR.} \textit{``without PU''} (brown): performance of XGBoost model without applying DEDPUL or PULSNAR to identify probable positives. \textit{``with DEDPUL''} (golden): performance of XGBoost model after identifying probable positives using DEDPUL. \textit{``with PULSNAR''} (blue): performance of XGBoost model after identifying probable positives using PULSNAR. The error bar represents a 95\% confidence interval of the metric.}
\label{fig:dry_beans_classification}
\end{figure}

\begin{figure}
\centering
\includegraphics[width=0.95\columnwidth]{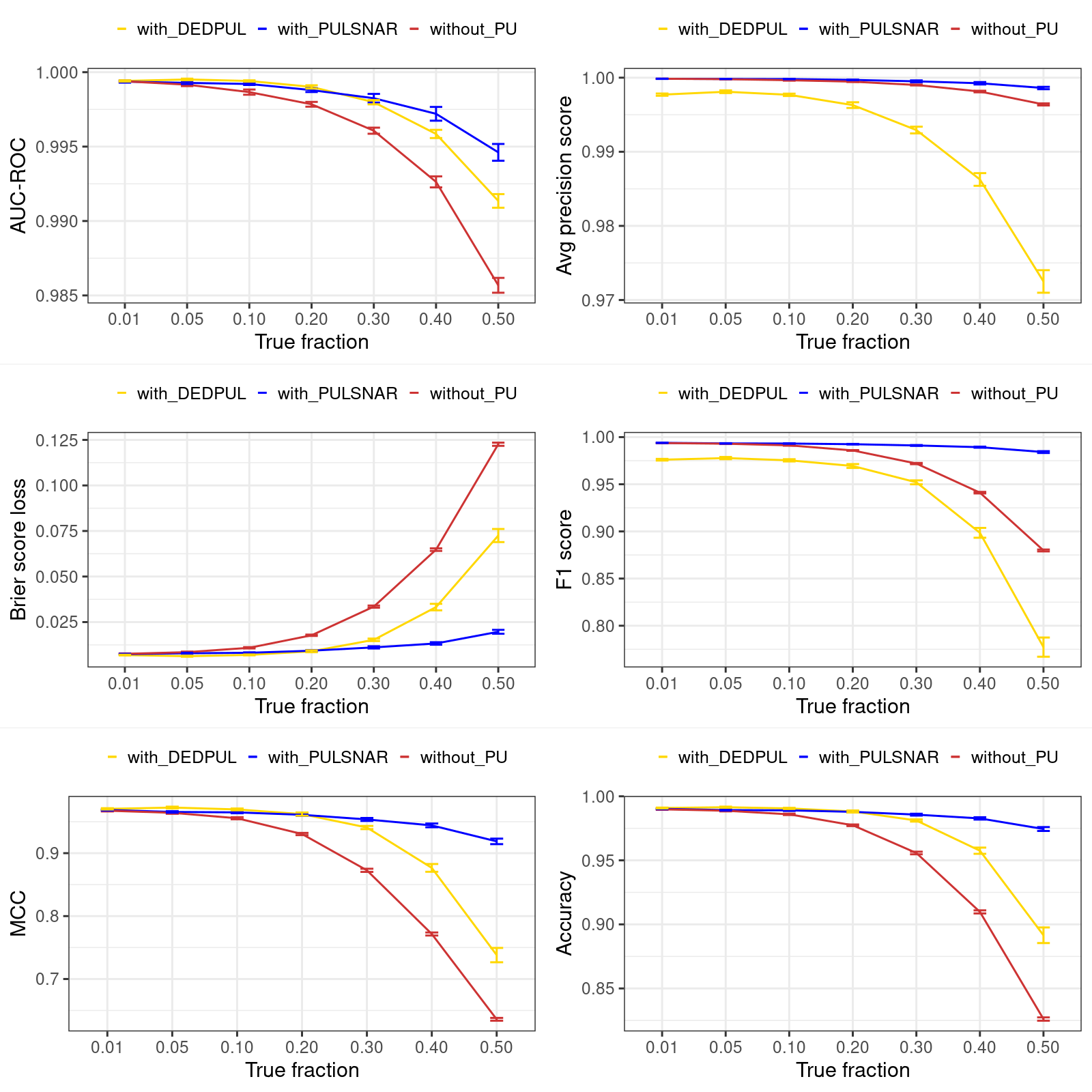}
\caption{\textbf{Classification performance of XGBoost model on smartphone SNAR dataset with/without DEDPUL and PULSNAR.} \textit{``without PU''} (brown): performance of XGBoost model without applying DEDPUL or PULSNAR to identify probable positives. \textit{``with DEDPUL''} (golden): performance of XGBoost model after identifying probable positives using DEDPUL. \textit{``with PULSNAR''} (blue): performance of XGBoost model after identifying probable positives using PULSNAR. The error bar represents a 95\% confidence interval of the metric.}
\label{fig:smartphones_classification}
\end{figure}

Figures \ref{fig:dry_beans_classification} and \ref{fig:smartphones_classification} show the classification performance of the XGBoost models with/without identifying probable positives using DEDPUL and PULSNAR. As the true fraction of positives among the unlabeled set increased for both datasets, the $\alpha$ estimates by DEDPUL were not close to true fractions, leading to a decrease in its classification performance. No significant drop in the classification performance of PULSNAR was observed, as the $\alpha$ estimates by PULSNAR were close to true answers. 

\end{document}